\documentclass[11pt]{article}
\newtheorem{theorem}{Theorem}% [section]

\newtheorem{proposition}[theorem]{Proposition}
\newtheorem{lemma}[theorem]{Lemma}

\newenvironment{proof}{\noindent {\bf Proof. }}{\hfill $\Box$ \\}
\newfont{\mfoo}{cmssdc10 scaled\magstep1}
\newfont{\mfo}{cmtt9 scaled\magstep1}

\usepackage{amsmath,amssymb}
\DeclareMathOperator*{\argmax}{arg\,max}

\DeclareMathOperator*\lowlim{\underline{lim}}
\DeclareMathOperator*\uplim{\overline{lim}}
\usepackage{xpatch}
\xpretocmd{\eqref}{Eq.~}{}{}
\usepackage[toc,page]{appendix}
\usepackage{MnSymbol}%
\usepackage{wasysym}
\usepackage[latin1]{inputenc}
 \usepackage{tgtermes}
\usepackage{appendix}
\usepackage{enumerate}
\usepackage{bbm}
\usepackage{tcolorbox}
\usepackage{adjustbox}
\usepackage{bm}
\usepackage{color}
\usepackage[left=4.00cm, right=3.00cm, top=4.00cm, bottom=3.00cm]{geometry}
\usepackage{graphicx}
\usepackage{pdfsync}
\usepackage{tikz}
\usepackage[active]{srcltx}
\usepackage{pgf}
\usepackage{mathtools}
\usetikzlibrary{shadows}
\usepackage{accents}
\usepackage{pdfsync}
\usepackage{soul}
\newfont{\csmall}{cmcsc10 scaled 900}
\newfont{\mbs}{cmmib7}
\newfont{\mbbs}{cmmib5}
\newcommand{\hsp}{\mbox{\ \ \ }} 
\newcommand*{\field}[1]{\mathbb{#1}}%

\newcommand{\bp}{\ensuremath{b^{'}}}
\newcommand{\tpi}{\ensuremath{\tilde{\pi}}}

 \newcommand{\cmdNext}[1][{}]{%
\setbox2=\hbox{$\scriptstyle{#1}$}\ifdim \wd2<\wd1\setbox2=\hbox to \wd1{\hfill$\scriptstyle{#1}$\hfill}\fi
\ifdim\wd2>\wd1\hspace*{.6\wd2}\fi
{\cmdDefaultA}_{\llap{$\box2$}}\!{\sf T}
}
\def\elabel#1{\label{#1}%
\vbox to 0pt{\vspace{.08cm}%
\hbox to 0pt{\hspace{-3cm}\tiny (e:#1)}\vss}}

\def\eq#1/{(\ref{#1})}
\def\beq{\begin{equation}}
\def\eeq{\end{equation}}
\catcode`\@=11
\@addtoreset{equation}{section}
\catcode`\@=12

\long\def\notes#1{\ifinner
             {\tiny #1}
             \else
              \marginpar{\protect\tiny #1}
              \fi}

\usepackage{endnotes}
\let\footnote=\endnote

\usepackage[round]{natbib}
 \newcommand{\Rew}{\mathcal{V}}
\newcommand{\CC}{\mathcal{C}}
\newcommand{\cc}{\citet}
\title{Optimal Data Driven  Resource Allocation under Multi-Armed Bandit Observations}
\author{Apostolos N. Burnetas\footnote{Department of Mathematics, University of Athens, Athens, Greece, E-mail: aburnetas@math.uoa.gr} \and  Odysseas Kanavetas\footnote{Mathematical Institute, Leiden University, Einsteinweg 55, Leiden, 2333 CC, The Netherlands,
E-mail: o.kanavetas@math.leidenuniv.nl} \and Michael N. Katehakis\footnote{Department of Management Science and Information Systems, Rutgers University, Piscataway, NJ 08854, USA,
E-mail: mnk@rutgers.edu}
}
\begin{document}
\maketitle
\date
%%%%%%%%%%%%%%%%
 
%\ARTICLEAUTHORS{%
%\AUTHOR{}
%\AFF{Department of Mathematics, National and Kapodistrian  University, Panepistemiopolis, Athens 15784, Greece, \EMAIL{aburnetas@math.uoa.gr}} %, \URL{}}
%\AUTHOR{Odysseas Kanavetas} \AFF{Department of Industrial
%Engineering, Sabanci University, Orhanli Tuzla, Istanbul 34956,
%Turkey,\\
% \EMAIL{okanavetas@sabanciuniv.edu}} \AUTHOR{Michael N.
%Katehakis} \AFF{Department of Management Science and Information
%Systems, Rutgers University, Piscataway, NJ 08854, USA,
%\EMAIL{mnk@rutgers.edu}}
%% Enter all authors
%} % end of the block

\begin{abstract}
This paper introduces the first 
 asymptotically optimal strategy for a multi armed
bandit (MAB) model under side constraints. The side constraints  model situations in which
 bandit  activations  are limited by the availability of certain  resources that are replenished at a constant rate.  
 The main result involves the derivation of an asymptotic lower bound for the regret of feasible uniformly fast policies and the construction of policies that achieve this lower bound, under pertinent conditions.    Further,  we  provide  the explicit
form of such policies for the case in which the
 unknown distributions   are Normal with unknown means and known variances, for the case of Normal distributions with unknown means and unknown variances and for the case of arbitrary discrete distributions with finite support. \\
 \vspace*{\baselineskip}\noindent

\vspace*{\baselineskip}\noindent
%{\csmall AMS Subject Classification} 
\end{abstract}

% Enter your abstract
%

%\KEYWORDS{deterministic inventory theory; infinite linear programming duality;
%  existence of optimal policies; semi-Markov decision process; cyclic schedule}


% Samples of sectioning (and labeling) in MNSC
% NOTE: (1) \section and \subsection do NOT end with a period
%       (2) \subsubsection and lower need end punctuation
%       (3) capitalization is as shown (title style).
%
%\section{Introduction.}\label{intro} %%1.
%\subsection{Duality and the Classical EOQ Problem.}\label{class-EOQ} %% 1.1.
%\subsection{Outline.}\label{outline1} %% 1.2.
%\subsubsection{Cyclic Schedules for the General Deterministic SMDP.}
%  \label{cyclic-schedules} %% 1.2.1
%\section{Problem Description.}\label{problemdescription} %% 2.

% Text of your paper here

\section{Introduction}

Consider the problem of sequentially  activating one of a finite number of
independent   bandits, where each activation of a bandit  incurs a number of bandit dependent  
resource utilizations, or activation costs. For each resource type the  constraint ensures that the total resource utilized (or equivalently  cost incurred) at any   time does not exceed the current resource availability (budget).   It is assumed that following each activation any unused resource  amounts can be carried forward for use in future activations.  We also make the  assumption that successive activations of  each 
bandit yield  independent, among different bandits,  identically distributed (iid) random rewards  with  positive means, and   distributions that depend on unknown parameters. 
The objective is to obtain  a feasible policy    that
maximizes asymptotically   the total  expected rewards or equivalently,  
  minimizes asymptotically a regret function.   
  We develop a class of   feasible policies that are shown to be asymptotically optimal within a large class of good policies that  
  uniformly fast (UF) convergent, in the sense of \cc{bkmab96} and \cc{Lai85}.
  The results in this paper  extend the work in  \cc{bkkmab17}  which solved the   case where there exists only one type of   constraint for all bandits. Further, the   class of block-UCB (b-UCB)  feasible  
policies developed here, which achieve the asymptotic lower bound in the regret,   have  a simpler form and are easier to compute than those in   \cc{bkkmab17}.   We also refer to
 \cc{bu+ka12}  where  a consistent  policy (i.e., with
regret $o(n)$) for the case of a single linear constraint was constructed.  %As far as we know, the    first formulationof the MAB problem with a side constraint considered herein was
%given  in \cc{bkmab98}. 

There is an extensive literature on the  multi-armed bandit (MAB)
problem, cf.
\cc{mahajan2008multi,audibert2009exploration,auer2010ucb,honda2011asymptotically,bubeck2012best,lattimore2018ci}, \cc{cowan18s}, 
and references therein.  
MAB models with a finite exploration budget  that  limits the number of times one can   sample (activate)  arms during  an initial exploration phase,   which is
    used to identify  the optimal arm  are considered in  \cc{bubeck2009pure}.  \cc{tran2010epsilon} considers the problem when both the exploration and exploitation phases are limited by a single budget and   establish an upper bound for the loss of a budgeted  $\epsilon$- first algorithm  for this problem.   
    \cc{badanidiyuru2018bandits} consider the  MAB problem with  multiple resource constraints and a finite horizon $T$, assuming that when a resource is exhausted activations stop.
    They show  how to construct policies with regret in the order of $O(\log T)$, where $T$ is the horizon length. 
    \cc{agrawal2014bandits} provided a    more  general version  of    \cc{badanidiyuru2018bandits}  which allows arbitrary concave objectives and convex feasibility constraints. \cc{ding2013multi} constructed UF policies (i.e.,  with regret
$O(log n)$) for cases in which activation costs are bandit-dependent iid random variables.     Applications of  MAB models   include  problems of online   revenue management: \cc{Simchi-Levi-online2018},  \cc{wang2014close}, \cc{johnson2015online}  of dynamic procurement: \cc{singla2013truthful},  auctions: \cc{tran2014efficient}.  %

One key difference between  these  finite  resource  budget models  and the model herein is that
in the present model is that we do not have a total  budget for each resource fixed at the beginning, but rather for each resource its budget  is increased at a constant rate at each activation period. Furthermore, the resource constraints must be satisfied at each   period in the sense that for each resource its  total budget utilization   during the first $n$ periods cannot exceed the total   
 $n$-period budget available for all $n$. Thus, for each resource there is  one constraint at each time period rather than a single constraint for the entire horizon.  In this way  the bandit activation problem becomes    more restricted and it requires  a different approach in the activation policies. For example, if a particular bandit consumes a large amount of some resource at each activation, then after one activation the controller  may have to wait for several subsequent periods until the budget of this resource is sufficiently replenished so that the bandit may be activated again. Thus the exploration phase is  necessarily intertwined with the exploitation phase due to the structure of the resource constraints.  
 
 %{\color{red} read this paragraph carefully and correct if necessary}.

%{\color{red} Note for us: They observe both reward and resource consumptions after each activation. I think only one bandit is activated each time. The difference is that their resources are depleted at random rates after each activation whereas in our case the resource consumptions by each bandit are fixed, and the resource budgets are increased at each period. This makes the problem harder in some sense, because we don't have the freedom to activate expensive bandits whenever we like, but we may have to wait for some resources to be replenished before activating them. This is the essence of constraint (3). I changed the discussion of the key differences below. Read carefully!! }

% any unused resource amounts can be carried over for use in future activations,  while maintaining feasibility as defined in terms of  the constraints of     \eqref{eq:ccon}.  As a result, bandits with large resource utilizations (cost coefficients) can  be activated if the controller does not utilize the full amount of resources in one (or more) period(s) so as to carry over the excess resources for use in future periods.

A second key difference is that we construct a new class of feasible UCB policies and we 
establish  their asymptotic optimality. Asymptotic optimality means that our policies achieve the exact asymptotic lower bound in the regret function and not only in terms of order of magnitude $O(\log n)$, as is typical in finite horizon formulations.  %We note that in most instances  asymptotically optimal policies have very good finite horizon performance.

On the applications side, the results herein can be used to solve infinite horizon versions of online network revenue management  where the retailer must price several unique products, each of which may consume common resources  (e.g., inventories of different products) that have limited availability and are replenished at a constant rate.
For versions of such problems with no resource (inventory)  replenishment we refer to  \cc{Simchi-Levi-online2018} and references therein.  Additional applications include  
  search-based and targeted advertising online learning, cf. \cc{rusmevichientong2006adaptive}, \cc{agarwal2014budget} and references therein.  

For other recent related work we
refer to: \cc{guha2007approximation,tran2012,thomaidou2012toward,lattimore2014optimal,sen2015adaptive,pike2017optimistic,zhou2018cost,spencer2018item} and 
  \cc{dena2013} \cc{cowan2015multi}  \cc{pike2018bandits},   \cc{lattimore2018bandit}, \cc{pike2017optimistic}, \cc{cowan23s}.
Similar action constrained  optimization problems also arise in MDPs  cf. \cc{feinberg1994constrained}, \cc{borkar2014risk},  queueing \cc{floske89}, 
  many areas c.f.  \cc{perakisnv}, \cc{perakis2015data}, \cc{btang2016},
\cc{feng-Sh2018}, \cc{kanav21}.

In the sequel we first establish in Theorem \ref{th:lb}, a necessary asymptotic lower bound for the rate of increase of the regret function of f-UF policies. We then construct a class of ``block f-UF'' policies and provide conditions under which they are asymptotically optimal within the class of f-UF policies, achieving this asymptotic lower bound, cf. Theorem \ref{th:ub}.  For the development of these policies we use the notion of `blocks of activations', that essentially  allow the implementation of necessary randomizations cf. \cc{feinberg1994constrained}, without violating the feasibility constraints.  Then,  in Section 4.1   we provide the explicit form of an asymptotically optimal f-UF policy for the case in which the unknown distributions are Normal with unknown means and known variances, in Section 4.2 for the case of Normal distributions with unknown means and unknown variances and in Section 4.3 we do the same for case where the unknown distributions are non parametric, discrete  with finite support.

\section{Model Formulation}
Consider $k$  independent
bandits, where successive
activations of a bandit $i,$ constitute a sequence of i.i.d. random
variables $X_{1}^{i},X_{2}^{i},\ldots$.  For each fixed $i,$  $X_{t}^{i},$ follows a univariate
distribution with density $f_{i}(\,|\underline{\theta}_i)$ with
respect to a nondegenerate measure $v$. The density $f_i (\,|\,)$ is
known and $\underline{\theta}_i$ is a vector of parameters belonging to some
set $\Theta_i$. Let
$\underline{\underline{\theta}}=(\underline{\theta}_1,\ldots,\underline{\theta}_k)$
denote the set of parameters,
$\underline{\underline{\theta}}\in\Theta$,  where
$\Theta\equiv\Theta_1\times\ldots\times\Theta_k$.
%Thus the   model is   determined by the vector $\underline{\underline{\theta}}$.
Given $\underline{\underline{\theta}}$ let
$\underline{\mu}(\underline{\underline{\theta}})=(\mu_1
(\underline{\theta}_1),\ldots,\mu_k (\underline{\theta}_k))$ be the
vector of the  expected values, i.e. $\mu_i
(\underline{\theta}_i)=E_{\underline{\underline{\theta}}}(X^i)$. The
true value  $\underline{\underline{\theta}}_0$  of
$\underline{\underline{\theta}}$ is unknown. We make the assumptions
that outcomes from different bandits  are independent and all means $\mu_i
(\underline{\theta}_i)$ are positive.

Each activation of bandit $i$ incurs $L$ different types of resource utilization (or cost): 
$c^i_{j}$, $j=1,\ldots,L.$ 
%, and without loss of generality
%for each type $j$, we assume $c^1_{j} \le c^2_{j} \le \ldots \le
%c^k_{j},$ and not all $c^i_{j}$ are equal. 
To avoid trivial cases, we will assume that $L<k.$ We also assume that there are two classes of bandits, one with $c^i_{j}< c^0_{j}$ for each constraint type $j$ and for any bandit $i$ in this class, and the other one with $c^i_{j}> c^0_{j}$ for each constraint type $j$ and for any bandit $i$ in this class. Intuitively, this means that there is at least one bandit that can be activated in each period without violating the strict feasibility constraint we define below. Now, by relabeling  the  bandits we call as bandit $i=1$   the bandit which has the maximum number of costs $c^1_j$  that are the minimum among  $c^i_j$   in the    constraint type $j$.  Similarly, we label  as bandit $k$   the bandit which has the maximum number of costs $c^k_j$  that are the maximum among  $c^i_j$   in the    constraint type $j$.  Note that   $c^1_{j}< c^0_{j}$ for each constraint type $j,$ and $c^0_j< c^k_{j}$,  again for each
 constraint  type $j$. 
For simplicity of the mathematical analysis below we assume that there is no bandit with activating cost $c^i_j$ that is equal to $c^0_j.$  Equivalently,  for each constraint type $j$, there exists   $d_j$, with  
  $1\leq d_j< k$ and $c^{d_j}_{j}< c^0_{j}<c^{d_j+1}_{j}$  (note that $d_j=\max\{ i\ :\ c^i_{j} < c^0_{j}\} $). 
  
Following standard terminology,  
adaptive policies   depend only on   past    
activations and observed outcomes. 
Specifically, let $A_t,X_t$ , $t=1,2,\ldots$ denote the bandit
activated and the observed outcome at period $t$. Let $h_t =(\alpha_1
,x_1,\ldots.,\alpha_{t-1},x_{t-1})$ denote a history of activations and
observations available at period t.
An adaptive policy is  a sequence $\pi =(\pi_1,\pi_2,\ldots)$ of
history dependent probability distributions on $\{1,\ldots,k\}$, such
that $ \pi_t (j,h_t)=P(A_t =j\,|\,h_t).$ Given $h_n$, let
$T^{\alpha}_{\pi}(n)$ denote the number of times bandit $\alpha$
has been activated during the first n periods  $T^{\alpha}_{\pi}(n) =
\sum_{t=1}^{n}1\{A_t =\alpha\} $. Let $\Rew_{\pi}(n)$ and
$\CC_{j,\pi}(n)$  be respectively the total reward earned and  total type $j$ resource utilized 
(cost incurred)  up to period $n$, i.e.,
\begin{equation}\label{eq:rew}
 \Rew_{\pi}(n)=
\sum_{i=1}^k     \sum_{t = 1}^{T^i_\pi(n)}  X^i_t,
\end{equation}
\begin{equation}\label{Eq:cost}
\CC_{j,\pi}(n)= \sum_{i=1}^k     \sum_{t = 1}^{T^i_\pi(n)}  c^i_{j} .
\end{equation}
 We call an adaptive  policy feasible if
\begin{equation}\label{eq:ccon}
 {\CC_{j,\pi}(n)}/{n} \leq c^0_{j}, \ \ \forall  j=1,\ldots,L, \ \forall n=1,2,\ldots
\end{equation}
The objective is to obtain  a feasible policy $\pi$     that
maximizes asymptotically  $E_{\underline{\underline{\theta}}}
\Rew_{\pi}(n),$ $ \forall \underline{\underline{\theta}}\in
\underline{\underline{\Theta}},$ or equivalently,  
  it  minimizes asymptotically the regret function
$R_{\pi}(\underline{\underline{\theta}},n)$ cf.  \eqref{reg1}.

\subsection{Optimal Solution Under Known  Parameters} 

It follows from standard theory of MDPs cf.  \cc{derman70},  that if  all parameters  
$\underline{\underline{\theta}}$ were known, the optimal activation(s) (the same in all periods) for maximizing  the expected average
reward are obtained as the solution to
 the following  linear program (LP).
\begin{eqnarray}
z^{*}(\underline{\underline{\theta}}) & = & \max
\sum_{i=1}^{k}\mu_{i}(\underline{\theta}_i) x_i \nonumber\\
subject \ to &&\sum_{i=1}^{k}c^i_{1} x_i  +y_1 =   c^0_{1} \nonumber \\
&& \hspace{1cm}
\vdots \label{lp}\\
&&\sum_{i=1}^{k}c^i_{L} x_i  +y_L =   c^0_{L} \nonumber \\
&&\sum_{i=1}^{k}x_i =1 \nonumber\\
&&x_i \geq 0,\forall i\ y_j \geq 0,\forall j,\nonumber
\end{eqnarray}
where the variables  $x_i$, for $i=1,\ldots,k$, represent  the activation probabilities 
for bandit $i$ of an optimal randomized policy.

Thus, a basic matrix $B$ is an $(L+1)\times(L+1)$ matrix that  consists of one or at most $L+1$ bandit (and slack)  variables $x_i$ (and $y_i$); recall that $L<k$. 
 Note that any  basic feasible
solution (BFS) corresponding to such a choice of the matrix $B$ is uniquely determined by the vector corresponding to the choice of basic bandit variables:  $b=\{i_1,\ldots,i_j\}$, $j=1,\ldots,L+1.$  For simplicity,  the sequel we will not distinguish between  $B$ and $b$,  since if one knows one he knows the other. 
Thus, the vector $b$ uniquely determines a 
 corresponding (possibly randomized) activation policy with randomization probabilities $x_{i_1},\ldots,x_{i_j}$, $j=1,\ldots,L+1$ in $b$.
%Note that in the case of degenerate BFS
%$b$, more than one matrices $B$ correspond to the same $b$.
We use $K$ to denote the set of  bandits corresponding to a feasible  choice of $b,$  for simplicity written as 
\begin{equation}
  K=\{ b \ : \ b=\{i_1,\ldots,i_j\}, \ j=1,\ldots,L+1  \}.\nonumber
\end{equation}
\noindent  Given our assumptions on the $ c^i_{j}s $, it follows that the feasible region of \eqref{lp} is nonempty and  bounded, hence $K$ corresponds to  a 
finite number of BFS. 

In the sequel it will be more convenient  to work with the  
dual problem DLP   stated below.

\begin{eqnarray*}
z^{*}_{D}(\underline{\underline{\theta}}) & = & \min \ c^0_{1}g_{1} +\ldots+c_{L}^{0}g_{L}+g_{L+1}\\
subject \ to  && c^1_{1}g_{1} +\ldots+c_{L}^{1}g_{L}+g_{L+1} \geq \mu_1 (\underline{\theta}_1) \\ &&\hspace{1cm}
\vdots \label{dlp}\\
&& c^k_{1}g_{1} +\ldots+c_{L}^{k}g_{L}+g_{L+1} \geq \mu_k (\underline{\theta}_k) \\\ &&  \ g_{j} \geq 0, \ j=1,\ldots,L, \ \ g_{L+1}\in {\Bbb R}.
\end{eqnarray*}

For a basic matrix $B$ of LP, we let $v^B= (g^B_{1},\ldots,g^B_{L}, g^B_{L+1})$ denote the dual
vector corresponding to $B$, i.e., $v^B = \mu_B
(\underline{\underline{\theta}}) B^{-1}$, where $\mu_B
(\underline{\underline{\theta}})$ contains the means of the bandits  given by the choice of  $B .$

A BFS is optimal if and only if  the reduced costs (dual slacks) for the corresponding basic matrix $B$ 
are all nonnegative, i.e., 
\begin{equation}
  \label{eq:phidef}
\phi^B_{\alpha} (\underline{\underline{\theta}}) \equiv c^{\alpha}_{1}
g^B_{1}+\ldots+c^{\alpha}_{L}
g^B_{L}  + g^B_{L+1} - \mu_{\alpha}(\underline{\theta}_{\alpha})
% = v^B
% \left (
%   \begin{array}{c}
%     c_{\alpha} \\ 0
%   \end{array}
%  \right ) - \mu_{\alpha}
  \geq 0, \ \alpha=1,\ldots, k.\nonumber
\end{equation}

%Note that if
%an optimal BFS is degenerate, then not all basic matrices
%corresponding to it are necessarily optimal.
%

Note that  it  is easy to show that the reduced cost can be expressed as a
linear combination of the   bandit means, i.e.,
$\phi_{\alpha}^{B}(\underline{\underline{\theta}})=\underline{w}_{\alpha}^{B}\underline{\mu}(\underline{\underline{\theta}})$,
where $\underline{w}_{\alpha}^{B}$ is an appropriately defined
vector that does not depend on
$\underline{\mu}(\underline{\underline{\theta}})$.

In the sequel we use the notation
$O^*(\underline{\underline{\theta}})$ to denote
  the set of choices of $b$ corresponding to  optimal solutions of the LP for a
vector $\underline{\mu}(\underline{\underline{\theta}})$, i.e., $
O^*(\underline{\underline{\theta}})=\{ b \in K : b \mbox{ corresponds
to an optimal BFS} \}. $

\section{Optimal Policies Under Unknown Parameters}

\subsection{The Regret Function}
In this subsection we consider the case in which  $\underline{\underline{\theta}}$ is unknown and define the regret $R_{\pi}(\underline{\underline{\theta}},n)$   of a policy $\pi$ as the finite horizon loss in expected reward with respect to the optimal policy $\pi^*$ corresponding to the case in which  $\underline{\underline{\theta}}$ is known, i.e., 
\begin{eqnarray}\label{reg1}
R_{\pi}(\underline{\underline{\theta}},n)&=&
nz^{*}(\underline{\underline{\theta}})-
E_{\underline{\underline{\theta}}} \Rew_{\pi}(n) \nonumber\\
&=&n z^*(\underline{\underline{\theta}})- \sum_{j=1}^{k}\mu_j
(\underline{\theta}_j)
E_{\underline{\underline{\theta}}}T_{\pi}^{j}(n).
\end{eqnarray}

We now state the following.  
 A  feasible policy $\pi$ is called consistent      if
$
R_{\pi}(\underline{\underline{\theta}},n)=o(n), \
n\rightarrow\infty, \    \forall \
\underline{\underline{\theta}}\in\Theta,
$ 
and it is called  uniformly fast (f-UF)  if
 $
R_{\pi}(\underline{\underline{\theta}},n)=o(n^{a}), \
n\rightarrow\infty, \ \forall \ a>0, \ \forall \
\underline{\underline{\theta}}\in\Theta. $

Following the approach in  \cc{bkkmab17}, we  will establish in Theorem \ref{th:lb} below a lower bound $M(\underline{\underline{\theta}})$  for   the regret of any 
  f-UF policy  and construct a block UCB policy $\pi^0 $ which is f-UF and its regret achieves this lower bound, i.e., 
   $$\lowlim_{n \rightarrow \infty}R_{\pi^0}(\underline{\underline{\theta}},n)/\log n \leq
M(\underline{\underline{\theta}})  , \ \forall \underline{\underline{\theta}}.$$
Thus,  it will be shown that  policy
$\pi^0$  is asymptotically optimal. 

\subsection{Lower Bound for the Regret}

For any optimal basis choice $b=\{i_1,\ldots,i_j\}  \in O^*(\underline{\underline{\theta}})$,   and $\alpha \in  b $, 
 we  define the sets  $\Delta\Theta_{\alpha}(\underline{\underline{\theta}})$  and $D(\underline{\underline{\theta}})$,
as follows
\begin{equation}\label{eq:delta}
\Delta\Theta_{\alpha}(\underline{\underline{\theta}})= \{
\underline{\theta}_{\alpha}^{'}\in\Theta_{\alpha}:
O^*(\underline{\underline{\theta}}^{'})=\{b\} \, \},
\end{equation}
\begin{equation}
D(\underline{\underline{\theta}})=\{ \alpha: \ \alpha \notin b
\mbox{ for any } b\in O^*(\underline{\underline{\theta}}) \mbox{ and }
\Delta\Theta_{\alpha}(\underline{\underline{\theta}})\neq\emptyset
\}, \nonumber
\end{equation}
where $\underline{\underline{\theta}}^{'}=(\underline{\theta}_1,
\ldots,\underline{\theta}_{\alpha}^{'},
\ldots,\underline{\theta}_k)$, is a new vector such that only
parameter $\underline{\theta}_{\alpha}^{'}$ is changed from
$\underline{\theta}_{\alpha}$.
Note that the first  set consists of  all   values of
$\Theta_{\alpha}$ under which the problem  with   known  parameters under
the perturbed $\underline{\underline{\theta}}^{'}$ has a unique
optimal solution that   includes bandit $\alpha$. The second set
$D(\underline{\underline{\theta}})$, consists of  all
  bandits that do not appear in any optimal solution
under parameter set $\underline{\underline{\theta}}$ but, by changing 
only the parameter vector  $\underline{\theta}_{\,\alpha}$, there is uniquely optimal
solution  that contains them. 

We next define the minimum distance   of a  parameter
 vector  $\underline{\theta}_{\alpha}$ to a new parameter vector   $\underline{\theta}_{\alpha}^{'}$ which makes 
bandit $\alpha$ to become optimal and hence appear in the unique optimal 
solution when its parameter becomes  $\underline{\theta}_{\alpha}^{'}$ .
\begin{eqnarray}\label{eq:k1}
K_{\alpha}(\underline{\underline{\theta}})&=& \inf\{
I(\underline{\theta}_{\alpha},\underline{\theta}_{\alpha}^{'}): \
\underline{\theta}_{\alpha}^{'}\in\Delta\Theta_{\alpha}
(\underline{\underline{\theta}}) \} , 
\end{eqnarray}
where,  $I(\underline{\theta}_{\alpha},\underline{\theta}_{\alpha}^{'})$
denotes the Kullback-Leibler distance  for the distributions $f(\cdot,\underline{\theta}_{\alpha}), $ and $f(\cdot,\underline{\theta}_{\alpha}^{'} )$ i.e.,  
\begin{equation}
I(\underline{\theta}_{\alpha},\underline{\theta}_{\alpha}^{'})=\int_{-\infty}^{+\infty}
\log
\frac{f(x;\underline{\theta}_{\alpha})}{f(x;\underline{\theta}_{\alpha}^{'})}
f(x;\underline{\theta}_{\alpha})dv(x).\nonumber
\end{equation}

The next Lemma establishes lower bounds for  the new mean 
$\mu_{\alpha}(\underline{\theta}_{\alpha}^{'})$ under the changed parameter vector $\underline{\theta}_{\alpha}^{'}$ in terms of the quantity 
$\mu_{\alpha}^*(\underline{\underline{\theta}})=\phi^B_{\alpha}
(\underline{\underline{\theta}})
+\mu_{\alpha}(\underline{\theta}_{\alpha})$. 
The proof  is specialized and not the focus of this paper,  and is relegated to the appendix.

\begin{lemma}\label{lem:bd1} For any  optimal matrix $B$ under $\underline{\underline{\theta}}$, such
that for any $\underline{\theta}_{\alpha}^{'}\in\Delta\Theta_{\alpha}(\underline{\underline{\theta}})$ the following is true  
\begin{align*}
 \phi^B_{j}
(\underline{\underline{\theta}}^{'})&=\phi^B_{j}(\underline{\underline{\theta}})\geq 0, \ \forall \ j\neq\alpha \mbox{,  and}\\
\phi^B_{\alpha}
(\underline{\underline{\theta}}^{'})&= \mu_{\alpha}^*(\underline{\underline{\theta}}) -\mu_{\alpha}(\underline{\theta}_{\alpha}^{'})<0.
\end{align*}

%
%\ \ \ $(ii)$
%$\mu_{\alpha}^*(\underline{\underline{\theta}})<\mu_{\alpha}(\underline{\theta}_{\alpha}^{'})<\mu_{\alpha}^*(\underline{\underline{\theta}})+\rho$,
%where $\rho>0$.
\end{lemma}

 The above and Eqs. (\ref{eq:delta}),  (\ref{eq:k1}) imply  that  
\begin{equation}\label{eq:k2}
K_{\alpha}(\underline{\underline{\theta}}) =
 \inf\{
I(\underline{\theta}_{\alpha},\underline{\theta}_{\alpha}^{'}): \
\underline{\theta}_{\alpha}^{'}\in\Theta_{\alpha}, \
\mu_{\alpha}^*(\underline{\underline{\theta}})<\mu_{\alpha}(\underline{\theta}_{\alpha}^{'})\}.
\end{equation}
%where $\rho=\rho(\underline{\underline{\theta}},\alpha,B)>0.$

In order to establish  a lower bound on the regret we   need to express it as:
\begin{equation}\label{reg2}
R_{\pi}(\underline{\underline{\theta}},n)=\sum_{j=1}^{k}\phi_{j}^{B}(\underline{\underline{\theta}})
E_{\underline{\underline{\theta}}}T_{\pi}^{j}(n)+\sum_{i=1}^{L}g^B_{i} \
\sum_{j=1}^{k}(c^0_{i} -
c^j_{i})E_{\underline{\underline{\theta}}}T_{\pi}^{j}(n), \mbox{ $\forall\ \underline{\underline{\theta}}\in\Theta$},
\end{equation}
and any optimal basic matrix  $B,$  where the above expression follows from the LP and DLP relations since $B$ is an optimal basis:
$z^*(\underline{\underline{\theta}})=\sum_{i=1}^{L}c^0_{i} g^{B}_{i}+g^{B}_{L+1}$ and $\phi^{B}_{i} (\underline{\underline{\theta}}) = \sum_{j=1}^{L} c^{i}_{j}
g^{B}_{j}  + g^{B}_{L+1} - \mu_{i}(\underline{\theta}_{i})$.

 Both terms of the right side of  \eqref{reg2} are nonnegative, the
first due to  optimality of $B$ and the second due to the  feasibility of $B$.
It follows that a necessary and sufficient condition for a
policy $\pi$ to be f-UF  is that for all 
$\underline{\underline{\theta}}\in\Theta$ and any optimal  $B$
under $\underline{\underline{\theta}}$ the following two relations hold.

\begin{equation}\label{ug1}
\phi_{j}^{B}(\underline{\underline{\theta}})\uplim_{n\rightarrow\infty}\frac{E_{\underline{\underline{\theta}}}T_{\pi}^{j}(n)}{n^a}=0,
\mbox{ for all }  a>0, \  j\notin b,
\end{equation}
\begin{equation}\label{ug2}
 \sum_{i=1}^{L}g^B_{i}\sum_{j\in b}(c^0_{i} -
c^j_{i}) \uplim_{n\rightarrow\infty}\frac{E_{\underline{\underline{\theta}}}T_{\pi}^{j}(n)}{n^a}=0  \mbox{ for all }   a>0.
\end{equation}

The following  lemma and proposition    are used  to establish in Lemma \ref{lem:lb} a lower bound for the activation frequencies of any f-UF policy. They readily imply the lower bound of such polices for the regret in  Theorem \ref{th:lb}.  The  proof of the lemma   is relegated to the appendix.

\begin{lemma}\label{lem:g}If there is a uniquely optimal   $b\in
O^*(\underline{\underline{\theta}})$. Then the following hold.

$(i)$ If
$b=\{i_1,\ldots,i_j\}$, $j=2,\ldots,L+1$,  then  $g^B_i >0$, for $j-1$ out of the $L$ resource   constraints, and equal to $0$ for the remaining  resource constraints.  

$(ii)$ When $b$ is a singleton, i.e.,  $b=\{i_1\}$,  then  $g^B_i =0$, for every resource constraint  $i=1,\ldots,L$, i.e., only $g^B_{L+1} $ is positive.  
\end{lemma}

%\proof{Proof of Lemma 2.}$(i)$ Let
%$\underline{\underline{\theta}}:$
%$O^*(\underline{\underline{\theta}})=\{ b \}$, then $g^B_i >0$, only for $j-1$ out of $L$ type of constraints because if $g^B_i =0$ for some of them we must have
%more
%than one solutions in the primal, which cannot occur because $b$ is uniquely optimal.\\
%$(ii)$ Let $\underline{\underline{\theta}}:$
%$O^*(\underline{\underline{\theta}})=\{ b \}$,
%then $g^B_i =0$, for all $i=1,\ldots,L$ from the dual solution and
%$\phi_{j}^{B}(\underline{\underline{\theta}})>0$ for all $j\neq
%i_1$.
%\Halmos
%\endproof

The next proposition establishes that    a f-UF 
policy  is such that $\forall \ \underline{\underline{\theta}}\in
\underline{\Theta}$, it must be true that the number
of activations from each bandit  $  \alpha\in
D(\underline{\underline{\theta}})$ are at least
$\beta_n$,  for some  sequence of positive constants $\beta_n
=o(n)$. Its proof is given in   the appendix.

\begin{proposition}\label{prop:lb} For any f-UF policy $\pi$ and for all
$\underline{\underline{\theta}}\in\Theta$ we have that for
$\alpha\in D(\underline{\underline{\theta}})$, any
$\underline{\underline{\theta}}^{'}\in\Delta(\underline{\underline{\theta}})$
and for all positive sequences:  $\beta_n=o(n)$ it is true that
\begin{equation}
P_{\underline{\underline{\theta}}^{'}}[T^{\alpha}_{\pi}(n)
<\beta_n]=o(n^{a-1}), \mbox{ for all } a>0.\nonumber
\end{equation}
\end{proposition}

The next Lemma follows using a change of measure from $\underline{\underline{\theta}}^{'}$
to $\underline{\underline{\theta}}$ as done   in 
  \cc{bkmab96} and in \cc{Lai85}.

\begin{lemma}\label{lem:lb} If
$P_{\underline{\underline{\theta}}^{'}}[T^{\alpha}_{\pi}(n)
<\beta_n]=o(n^{a-1}),$ for all $a>0$ and a positive sequence
$\beta_n=o(n)$ then
\begin{equation}
\lim_{n\rightarrow\infty}P_{\underline{\underline{\theta}}}[T^{\alpha}_{\pi}(n)<\frac{\log
n}{K_{\alpha}(\underline{\underline{\theta}})}]=0,\nonumber
\end{equation}
for all $\underline{\underline{\theta}}\in\Theta$ and
$\alpha\in D(\underline{\underline{\theta}})$.
\end{lemma}
\proof{%Proof of Lemma \ref{lem:lb}.} 
 If we take $\beta_n={\log
n}/{K_{\alpha}(\underline{\underline{\theta}})}$ then Proposition \ref{prop:lb} implies 
$P_{\underline{\underline{\theta}}^{'}}[T^{\alpha}_{\pi}(n)
< {\log n}/{K_{\alpha}(\underline{\underline{\theta}})}]=o(n^{a-1}).$ 
Now, using the  change of measure from $\underline{\underline{\theta}}^{'}$
to $\underline{\underline{\theta}}$  and  the same arguments as in 
  \cite{bkmab96}  we have that
\begin{equation}
\lim_{n\rightarrow\infty}P_{\underline{\underline{\theta}}}[T^{\alpha}_{\pi}(n)<\frac{\log
n}{K_{\alpha}(\underline{\underline{\theta}})}] =0.\nonumber
\end{equation}
\endproof

%\proof{Proof of Lemma 3.}If we take $\beta_n=\frac{\log
%n}{K_{\alpha}(\underline{\underline{\theta}})}$ then
%$P_{\underline{\underline{\theta}}^{'}}[T^{\alpha}_{\pi}(n)
%<\frac{\log
%n}{K_{\alpha}(\underline{\underline{\theta}})}]=o(n^{a-1})$ and
%using a change of measure from $\underline{\underline{\theta}}^{'}$
%to $\underline{\underline{\theta}}$ and following the arguments in
%  \cite{bkmab96,lai85}  we have that
%\begin{equation}
%\lim_{n\rightarrow\infty}P_{\underline{\underline{\theta}}}[T^{\alpha}_{\pi}(n)<\frac{\log
%n}{K_{\alpha}(\underline{\underline{\theta}})}] =0.\nonumber\Halmos
%\end{equation}
%
%\endproof

We next define the constant $M(\underline{\underline{\theta}}) $ and
prove the main theorem of this section. Let
$$
M(\underline{\underline{\theta}}) =\sum_{j\in
D(\underline{\underline{\theta}})}\frac{\phi_{j}^{B}(\underline{\underline{\theta}})}{K_j(\underline{\underline{\theta}})}.
$$

\begin{theorem}\label{th:lb} If $\pi$ is an f-UF policy then
\begin{equation}
\lowlim_{n\rightarrow\infty}\frac{R_{\pi}(\underline{\underline{\theta}},n)}{\log
n}\geq M(\underline{\underline{\theta}}), \ \forall
\underline{\underline{\theta}}\in\Theta.\nonumber
\end{equation}
\end{theorem}

\proof{%Proof of Theorem \ref{th:lb}.} 
By Lemma \ref{lem:lb} and using the Markov inequality, we obtain that if $\pi$
is f-UF, then
\begin{equation}
\lowlim_{n\rightarrow \infty}\frac{E_{\underline{\underline{\theta}}}T^{j}_{\pi}(n)}{\log
n}\geq\frac{1}{K_{j}(\underline{\underline{\theta}})}, \  \forall
j\in D(\underline{\underline{\theta}}), \ \forall
\underline{\underline{\theta}}\in\Theta.\nonumber
\end{equation}

Also, we have from Lemma 2 that $g^{B}_{i}\geq0$ and from Eq.
(\ref{eq:ccon}), we have  that $nc^0_{i}-E_{\underline{\underline{\theta}}}C_{i,\pi}(n)\geq 0$, for all $n,i$.
Finally, we have that the optimal bandits under
$\underline{\underline{\theta}}$ have
$\phi_{j}^{B}(\underline{\underline{\theta}})=0$. 
These observations together with the above two relations  suffice to complete the proof if we recall that 
 $R_{\pi}(\underline{\underline{\theta}},n)=\sum_{j=1}^{k}\phi_{j}^{B}(\underline{\underline{\theta}})
E_{\underline{\underline{\theta}}}T_{\pi}^{j}(n)+\sum_{i=1}^{L}g^B_{i} \
\sum_{j=1}^{k}(c^0_{i} -
c^j_{i})E_{\underline{\underline{\theta}}}T_{\pi}^{j}(n).  $ 
%\begin{equation}
%\lowlim_{n\rightarrow\infty}\frac{R_{\pi}(\underline{\underline{\theta}},n)}{\log
%n}\geq\sum_{j\in
%D(\underline{\underline{\theta}})}\frac{\phi_{j}^{B}(\underline{\underline{\theta}})}{K_j(\underline{\underline{\theta}})},
%\mbox{ for all } \underline{\underline{\theta}}\in\Theta.\nonumber 
%\end{equation}
\endproof

\subsection{Blocks and Block Based Policies}
We consider a class of policies such that activation  is performed
in groups of   periods called {\it activation blocks} as defined below so as 
 total resource utilization of activations in each block satisfies all the resource  constraints of \eqref{eq:ccon} . 
For  each constraint $j$ we first define the
differences 
\begin{equation}\label{Eq:delta}
\delta^i_{j} \equiv c^i_{j} -c^0_{j}.\nonumber
\end{equation}
Note that $\delta^i_{j}$ expresses the net effect of a single activation of 
bandit $i$ on the corresponding resource $c^{0}_{j}$. This effect is a reduction in the resource $c$  if
$\delta^i_{j}>0,$ and  a surplus  if $\delta^i_{j}<0,$  that can be carried over to subsequent periods.

Thus, for any period the feasibility constraint of \eqref{eq:ccon} can be written as 
\begin{equation} \frac{1}{n}\sum_{t=1}^{n}\delta^{A_t}_{j} \leq 0, \
\forall \ n, \ j=1,\ldots,L.\nonumber
\end{equation}

Since $\delta^i_{j} $ is assumed to be rational, for each $i=1,\ldots,k$ and $j=1,\ldots,L$
and there is a finite number of them we may assume, without loss of
generality, that they are all integers, since we have assumed the same for the coefficients and right sides of the constraints. 

 Let $\mathcal{N}=\{ 1,\ldots,k \}$ be the set of all bandits. 
Intuitively,  for a specific   constraint 
``low resource utilization'' bandits in $\mathcal{N}$  (i.e., bandits with small $c^i_j$) 
must be sampled often enough to accumulate   resources (by carrying over `surplusess') in order to make possible the activation of 
``high resource utilization''  bandits. Mathematically it
suffices to find $\{ y^i_{j} , i\in \mathcal{N} \}$ such that bandit $i=1$ 
(where the $\delta^{1}_{j}=\min_{i}\delta^{i}_{j}$ for the maximum number of different constraint types $j=1,\ldots,L$)  
is sampled $y^{1}_{j}$ times and each bandit $i\in \mathcal{N}\setminus\{1\}$ is sampled once ($y^{i}_{j}=1$, for all $i\in \mathcal{N}\setminus\{1\}$ and $j=1,\ldots,L$), and $\sum_{i\in \mathcal{N}} y^i_j \delta^i_j \leq
0$, $y^i_j \in \field{N},$ $\forall \ i \in \mathcal{N}$ and $j=1,\ldots.,L$. Any block with $y^1_j=y^1_*=\max_{j}y^1_j$ for all $j=1,\ldots,L$
satisfying the previous properties is feasible with respect to the constraints of \eqref{eq:ccon}. 

%One
%possibility is to consider the smallest block, which will be
%appropriate in the incomplete information case. Thus the minimum
%length of the activating block, $\ell(\mathcal{N}),$ is the solution of the
%following linear program
%$$
%\ell(\mathcal{N}) =  \min \{ \sum_{i\in \mathcal{N}} m^i_j  \ : \ \sum_{i\in \mathcal{N}}m^i_j
%\delta^i_j \leq 0\ \& \ m^i_j \in \textbf{N}, \ \forall \ i\in \mathcal{N}, \ \ j=1,\ldots,J \}.
%$$

%
%\begin{eqnarray}
%\ell = & \min & \sum_{j\in J} m_j \nonumber\\
%&&\sum_{j\in J}m_j \delta_j \leq 0\nonumber\\
%&&m_j \in \textbf{N}, \ \forall \ j\in J.\nonumber
%\end{eqnarray}

%An optimal solution of LP specifies randomization probabilities
%that guarantee maximization of the average reward subject to the
%  constraints of the LP. The set of bandits into this optimal solution define
%an observable block can be    defined in (b) below.
%

Using the above remarks, we next define  the  Initial Sampling Block (ISB) and the Linear Programming
  Blocks (LPBs)  to construct  a class of block feasible policies $\mathcal{B} =\{\tpi\ : \tpi \ \mbox{is feasible} \}$  as follows.

{\bf ISB block:} A policy $\tpi\in \mathcal{B} $ starts with an ISB block during which
all bandits $\{ 1,\ldots,k \}$ are sampled at least a predetermined
number $n_0$ of times,  while the   constraints of \eqref{eq:ccon} are satisfied sample path-wise.
% with  a sufficient    number of samples taken  from the cheapest (smallest coefficient $c^{1}_{j}$ for each $j=1,\ldots,L$) bandit,
%so that the   constraints of \eqref{eq:ccon} are satisfied sample path-wise.  
This block is necessary in order to  obtain initial
 estimates $\mu_j (\underline{\hat{\theta}}_j)$ of
$\mu_j (\underline{\theta}_j)$ for all bandits. This 
  ISB block has     length of $y^1_*+(k-1)$,  defined above.

{\bf LPB blocks:} After the completion of an ISB block the   $\tilde{\pi}$  policy
chooses any $b$  (where $b= \{ i_1 \}$ or $b= \{ i_1,\ldots,i_j \}$)  that corresponds to a   BFS of the LP   and continues activating any of the  bandits in $b$ for a block of time
periods LPB($b$),  where LPB($b$)  is defined  as  follows.

\hsp i) When $b=\{ i_1 \}$. In this case    due to the feasibility of $b$ we must have $c^{i_1}_j < c^0_j$, for all $j=1,\ldots,L,$ and its activating
  frequency must  be equal to $1$, i.e.,  $x_{i_1}=1$. 
In this case we define the LPB($b$)  block to have  length equal to  $1$    and $\tilde{\pi}$  activates  bandit $i_1$   once, i.e.,  $m_{i_1}^{b}=1.$

\hsp  ii) When  $b=\{ i_1,\ldots,i_j \}$,   we have the positive 
randomization probabilities $\{x_l\}_{l\in b}.$ According to our assumption for rational coefficients  in the resource constraints, these randomization probabilities can be written in the form 
of  $x_{l}=\frac{m_{l}^{b}}{X(b)}, $ where $X(b) =\sum_{i=i_1}^{i_j}m_{i_j}^{b}$ is the least common denominator, for integer  numbers $m_{i_j}^{b}$ which we take to be the 
  number  of activations   of bandit $i_j$ within  this LPB$(b) .$ 
In this case we take  the length of the LPB($b$)    block to be equal to:
$\sum_{i=i_1}^{i_j}m_{i_j}^{b}$, and   we take \tpi \  to activate $m_{l}^{b}$ number of times
each bandit  $l \in b ,$ in this way ensuring  the constraint feasibility of
\tpi \  within  the block.

\noindent{\bf Remark 1}  In the above definition of block activations for  any $b$ that corresponds to a   BFS of the LP  we defined the integers  
  $m_{\alpha}^{b}$  to be  the number  of activations  from bandit $\alpha$ within  a LPB$(b).$ Note that in a computational implementation  the solution of an LP may be given in decimals. In this case, one cannot compute an  exact least common denominator for the randomization probabilities which is important since the denominator defines the length of the LP block.   However, every time one solves a LP one knows the specific constraint equations that correspond to the optimal basic matrix $B$.   Thus one has  a subsystem of equations of  the form $Bx=c$. Using the determinants expression of the solution  one   can compute an integer  denominator for the  randomization probabilities, under the assumption of rational coefficients.  Then one can find 
the least common denominator that is an integer and can be used as the length of the block of activations $\sum_{i=i_1}^{i_j}m_{i_j}^{b}$ corresponding to $B$. 
%We  note that we follow this procedure only for the LP which we decided to employ according to our index policy and not for every LP in the computational implementation.      

The definition of any $\tpi \in \mathcal{B} $  policy is completed by continuing  activations of bandits in $\mathcal{N}$ by repeating choices of  $b$  as above. 
In the sequel the choices of $b$ will be based on all collected data up to the start of the `current block' and thus $\tpi \in \mathcal{B}$ will be well defined adaptive policies. In what follows we 
will restrict attention to such policies in $\mathcal{B} $  and 
for notational simplicity we will simply write $\pi$ in place of
$\tpi$, when there is no risk for confusion.

%In the sequel the main difference in this analysis with that of \cc{bkkmab17}, is that now $b$ can contain more than two bandits and as a consequence the blocks have a more complex structure. 
 
\subsection{Regret of Block Based Policies}\label{sec:breg}

In this section we define the regret of block based policies and establish its relation with the initial regret of \eqref{reg1}.  
Assume that we have $l$ successive blocks  we take
$\widetilde{T}_{\pi}^b(l)$ to be  the number of LPB($b$) type blocks
  in first $l\geq 2$ blocks (since   for $l=1 $ we start with an ISB block).
Thus in the first $2,\ldots, l$ blocks each corresponds to a single feasible $b$ and we can write  $\sum_{b\in K}\widetilde{T}_{\pi}^b(l)=l-1$. 

Let $S_{\pi}(l)$
be the total length of first $l$ blocks (including the ISB block) and let $L_n=L_{\tpi}(n)$
denote the number of completed blocks in $n$ periods. It  can be  shown that
\begin{equation}\label{tblock}
T_\pi^\alpha(S_\pi(l))=\sum_{b\in K \, :\alpha\in
b}m_{\alpha}^{b}\,\widetilde{T}_\pi^b(l)+m^0_{\alpha},
\end{equation}
where $m_{\alpha}^{b}$ was defined above  and $m^0_{\alpha}$ is the number of
activations of bandit $\alpha$ in the ISB block (i.e., $m_1=y^1_*$ and $m^0_{\alpha}=m^0_{\alpha}1$ for all $\alpha\in \mathcal{N}\setminus \{1\}$). 

Note that when $\underline{\underline{\theta}}$ is known the quantity  
$E_{\underline{\underline{\theta}}}S_\pi(l)\,
z^{*}(\underline{\underline{\theta}})$ represents the total expected reward under an optimal policy.  When $\underline{\underline{\theta}}$ is unknown the quantity  
$E_{\underline{\underline{\theta}}}\sum_{j=1}^{k}\sum_{b\in
K}\mu_{j}(\underline{\theta}_j)\, m_{j}^{b}\,\widetilde{T}_\pi^b(l) \
+\sum_{j=1}^{k} \mu_{j}(\underline{\theta}_j)m^0_{j}$ represents the total expected reward under a block policy $\pi.$ Thus, we 
  can define the regret of a block policy $\pi$ as 
\begin{equation}\label{eq:regb}
\widetilde{R}_{\pi}(\underline{\underline{\theta}},l)=
E_{\underline{\underline{\theta}}}S_\pi(l)\,
z^{*}(\underline{\underline{\theta}})-
E_{\underline{\underline{\theta}}}\sum_{j=1}^{k}\sum_{b\in
K: j \in b}\mu_{j}(\underline{\theta}_j)\, m_{j}^{b}\,\widetilde{T}_\pi^b(l) \
-\sum_{j=1}^{k} \mu_{j}(\underline{\theta}_j)m^0_{j}. 
\end{equation}
Also note that in a period $n$ the length of the completed blocks $S_\pi(L_n))$ is less than or equal to $n$. When $S_\pi(L_n))=n$, then 
the number of activations of bandit $\alpha$ up to period $n$ is equal to the number of activations up to the last completed block, i.e., 
$T_\pi^\alpha(S_\pi(L_n)) = T^{\alpha}_{\pi}(n) .$ Otherwise, if $S_\pi(L_n))<n$  (i.e., period $n$ is within the last block which is uncompleted) then $T_\pi^\alpha(S_\pi(L_n)) <T^{\alpha}_{\pi}(n) .$  Note that there is a finite constant 
$M_{\alpha}$ that is equal to  the maximum number of times that  bandit $\alpha$ appears in every feasible  block (i.e., feasible basis).  This number allows one to obtain an upper bound on 
on $T^{\alpha}_{\pi}(n) $ when $S_\pi(L_n))<n,$  i.e., $ T^{\alpha}_{\pi}(n)  \leq
T_\pi^\alpha(S_\pi(L_n))+M_{\alpha}.$ Summarizing the above arguments we have:
\begin{equation}\label{eq:as3}
T_\pi^\alpha(S_\pi(L_n)) \leq T^{\alpha}_{\pi}(n)  \leq
T_\pi^\alpha(S_\pi(L_n))+M_{\alpha},
\end{equation}
 The definition of \eqref{eq:regb} and \eqref{eq:as3} yield the following  
relation for the two types of regret,
\begin{eqnarray}
&&\mbox{\ \ }\widetilde{R}_{\pi}(\underline{\underline{\theta}},L_n)
+(n-E_{\underline{\underline{\theta}}}S_\pi(L_n))\, z^*
(\underline{\underline{\theta}})
- \sum_{j=1}^{k}M_j \, \mu_{j}(\underline{\theta}_j)
\leq R_{\pi}(\underline{\underline{\theta}},n) 
\nonumber\\
&&\leq
\widetilde{R}_{\pi}(\underline{\underline{\theta}},L_n) +
(n-E_{\underline{\underline{\theta}}}S_\pi(L_n))\, z^*
(\underline{\underline{\theta}}).\label{as1}
\end{eqnarray}
The relations in  \eqref{as1} imply the following relation between the
two regret functions,
\begin{equation}
\uplim_{n\rightarrow\infty}\frac{R_{\pi}(\underline{\underline{\theta}},n)}{\log
n}
=\uplim_{n\rightarrow\infty}\frac{\widetilde{R}_{\pi}(\underline{\underline{\theta}},L_n)}{\log
L_n}.\label{as2}
\end{equation}
From \eqref{as2}, it follows that to show that 
 a policy   achieves the
lower bound for $R_{\pi}(\underline{\underline{\theta}},n)$ it
suffices to   show that it   achieves the lower bound for
$\widetilde{R}_{\pi}(\underline{\underline{\theta}},L_n)$.

\subsection{Asymptotically  Optimal Block UCB Policy}
In this section we provide a general method to construct
asymptotically optimal policies $\pi^0$. We call them Z-UCB policies and such policies achieve the lower bound
for the regret. To state the Z-UCB policy below we need some definitions. At the beginning of any block $l$ we have the estimates $\hat{\underline{\underline{\theta}}}^{l}$ which give an optimal solution $b(\hat{\underline{\underline{\theta}}}^{l})$  and corresponding optimal basis
 matrix $\hat{B}$   of LP$(\hat{\underline{\underline{\theta}}}^{l})$. Using the optimal solution of the linear program $b(\hat{\underline{\underline{\theta}}}^{l})$, we can    compute for  any block $l$ and for every bandit $\alpha \in \{1,\ldots,k\}$  the inflations 
$v_{\alpha} =v_{\alpha}(\hat{\underline{\underline{\theta}}}^l)$ of $\mu_{\alpha}(\underline{\hat{\theta}}_{\alpha}^{l})$ and  the set $\Phi_{l}^{(\hat{B},\hat{\underline{\underline{{\theta}}}}^{l})}$ of all  bandits $\alpha$ that according to Lemma 1, after  the
inflation of the mean only of bandit $\alpha$  to $v_\alpha$ may be in an optimal solution, as follows. 
\begin{align}
\ &  v_{\alpha}(\hat{\underline{\underline{\theta}}}^l)=\sup_{\underline{\theta}_{\alpha}^{'}}\{
\mu_{\alpha}(\underline{\theta}_{\alpha}^{'})\,:\,I(\hat{\underline{\theta}}_{\alpha}^{l},
\underline{\theta}_{\alpha}^{'})\leq\frac{\log
S_\pi(l-1)}{T_\pi^\alpha(S_\pi(l-1))} \},\label{vi}\\
\ & \Phi_{l}^{(\hat{B},\hat{\underline{\underline{\theta}}}^{l})}
=\{\alpha \ : \mu_{\alpha}^*(\hat{\underline{\underline{\theta}}}^{l})=\phi^B_{\alpha}
(\hat{\underline{\underline{\theta}}})
+\mu_{\alpha}(\hat{\underline{\theta}}_{\alpha})
< v_{\alpha}(\hat{\underline{\underline{\theta}}}^l)
\}.\label{phi}
\end{align}
 
 If
$\Phi_{l}^{(\hat{B},\hat{\underline{\underline{\theta}}}^{l})}\neq\emptyset$,
for every $\alpha \in
\Phi_{l}^{(\hat{B},\hat{\underline{\underline{\theta}}}^{l})}$ we
define the index: $u_{\alpha}(\hat{\underline{\underline{\theta}}}^{l})=u_{\alpha}(\hat{\underline{\underline{\theta}}}^{l},s,t )$ as follows.
\begin{equation}\label{eq:zucb}
u_{\alpha}(\hat{\underline{\underline{\theta}}}^{l},s,t ) \, = \,
\max_{\underline{\theta}_{\alpha}^{'}}\{
z^{b_{\alpha}(\hat{\underline{\underline{\theta}}}^{l},\underline{\theta}_{\alpha}^{'})}:
I(\hat{\underline{\theta}}_{\alpha}^{l},\underline{\theta}_{\alpha}^{'})\leq\frac{\log
s}{t} \}, 
\end{equation}
%
%%
%
%\begin{equation}
%u_{\alpha}(\hat{\underline{\underline{\theta}}}^{l},s,t )=
%\max_{\underline{\theta}_{\alpha}^{'}}\{
%z^{b_{\alpha}(\hat{\underline{\underline{\theta}}}^{l},\underline{\theta}_{\alpha}^{'})}:
%I(\hat{\underline{\theta}}_{\alpha}^{l},\underline{\theta}_{\alpha}^{'})\leq\frac{\log
%S_\pi(l-1)}{T_\pi^{\alpha}(S_\pi(l-1))} \},\nonumber
%\end{equation}
where $b_{\alpha}(\hat{\underline{\underline{\theta}}}^{l},\underline{\theta}_{\alpha}^{'})$ is the uniquely optimal solution 
 of LP$(\hat{\underline{\underline{\theta}}}^{l},\underline{\theta}_{\alpha}^{'})$, which is obtained from 
 the LP$(\hat{\underline{\underline{\theta}}}^{l})$ 
when  we replace only the parameter of bandit $\alpha$ by $\underline{\theta}_{\alpha}^{'}$.

We next state the asymptotically optimal {Z-UCB policy}.   It is based on the computation of the quantities 
$v_{\alpha}(\hat{\underline{\underline{\theta}}}^l),$
$\Phi_{l}^{(\hat{B},\hat{\underline{\underline{\theta}}}^{l})}$ and 
$u_{\alpha}(\hat{\underline{\underline{\theta}}}^{l},s,t )$
above, where $\hat{\underline{\underline{\theta}}}^l$ and $\hat{B}=\hat{B}(\hat{\underline{\underline{\theta}}}^{l}),$ are updated at the end of each block $l$. 
The update for $v_{\alpha}(\hat{\underline{\underline{\theta}}}^l)$  requires in general the solution of simple optimization problem with the single convex constraint and objective that depends on the functional relation of $\mu_{\alpha}( {\underline{\theta}}_{\alpha})$  on $ {\underline{\theta}}_{\alpha}.$
The updates for  $\Phi_{l}^{(\hat{B},\hat{\underline{\underline{\theta}}}^{l})}$  and $u_{\alpha}(\hat{\underline{\underline{\theta}}}^{l},s,t )$ are simple and fast since they  do not require solving any complex optimization problems.

\renewcommand{\thefootnote}{\alph{footnote}}. %mnk

\newcommand{\astfootnote}[1]{
\let\oldthefootnote=\thefootnote
\setcounter{footnote}{0}
\renewcommand{\thefootnote}{\fnsymbol{footnote}}
\footnote{#1}
\let\thefootnote=\oldthefootnote
}
%We can now state the asymptotically optimal {Z-UCB policy}.   

%\renewcommand{\thefootnote}{\fnsymbol{footnote}} % \renewcommand*{\thefootnote}{\fnsymbol{footnote}}
\begin{tcolorbox}[colback=blue!1, arc=3pt, width=.94\linewidth]
\textbf{Z-UCB POLICY $\pi^0$:}

{\bf Step 1}
Employ one ISB block  in 
order to have   one estimate $\hat{\underline{\theta}}_a$ from each bandit $a$. Then, update the vector of estimates 
 $\hat{\underline{\underline{\theta}}}^{2},$ and the statistics\footnote[1]{$S_{\pi^0}(l)$ is the total length of first $l$ blocks (including the ISB block) and $T_{\pi^0}^{\alpha}(S_{\pi^0}(l))$ is the number of times bandit $\alpha$
has been activated during the first $S_{\pi^0}(l)$ periods. Update them according to their definition in section \ref{sec:breg}.} $S_{\pi^0}(1)$ and $T_{\pi^0}^{\alpha}(S_{\pi^0}(1))$.\\
 
 {\bf Step 2} For block $l$ $(l>1)$, employ  an LPB($\pi^{0}(\hat{\underline{\underline{\theta}}}^{l})$) block, defined below. %according to 
 %a LPB($b^l$)  which corresponds to a feasible (not necessarily optimal solution $b^l=b^l(\hat{\underline{\underline{\theta}}}^{l})$ which is equal to 

 Given the history until the beginning of block $l$ we have the vector of estimates $\hat{\underline{\underline{\theta}}}^{l}$ and the statistics $S_{\pi^0}(l-1)$ and $T_{\pi^0}^{\alpha}(S_{\pi^0}(l-1))$. Based on these, compute\footnote[2]{The optimal policy of LP$(\hat{\underline{\underline{\theta}}}^{l})$ , the corresponding value of the objective function respectively.}
 $b^l_0$, and $z^{b^l_0}(\hat{\underline{\underline{\theta}}}^{l})$. Then compute  $v_{\alpha}(\hat{\underline{\underline{\theta}}}^l)$'s by Eq. (\ref{vi})  for every bandit $\alpha=\{ 1,\ldots,k \}$, and  $\Phi_{l}^{(\hat{B},\hat{\underline{\underline{\theta}}}^{l})}$,  by  Eq.~(\ref{phi}).\\
 %
%\end{tcolorbox}
%\begin{tcolorbox}[colback=blue!1, arc=3pt, width=.94\linewidth]
%\textbf{Z-UCB POLICY $\pi^0$ - Continued} \\
%

Now there are two cases:  

($i$)  If
$\Phi_{l}^{(\hat{B},\hat{\underline{\underline{\theta}}}^{l})}\neq\emptyset$,
for every $\alpha \in
\Phi_{l}^{(\hat{B},\hat{\underline{\underline{\theta}}}^{l})}$  compute
 the indices
$$u_{\alpha}(\hat{\underline{\underline{\theta}}}^{l})=u_{\alpha}(\hat{\underline{\underline{\theta}}}^{l},s,t )=u_{\alpha}(\hat{\underline{\underline{\theta}}}^{l},S_{\pi^0}(l-1),T_{\pi^0}^{\alpha}(S_{\pi^0}(l-1)) ),$$
 and the corresponding   uniquely optimal BFS:   
 $b^0_{\alpha}(\hat{\underline{\underline{\theta}}}^{l})=b_{\alpha}(\hat{\underline{\underline{\theta}}}^{l},\underline{\theta}_{\alpha}^0(s,t))$
 in Eq. (\ref{eq:zucb}).  %cf. discussion in page 5. 

 The {Z-UCB policy $\pi^0$} employs the LPB block which corresponds to the index:
\begin{equation}%\label{eq:pio}
\pi^{0}(\hat{\underline{\underline{\theta}}}^{l})=\argmax_{b^0_{\alpha}(\hat{\underline{\underline{\theta}}}^{l}),\ \alpha\in\Phi_{l}^{(\hat{B},\hat{\underline{\underline{\theta}}}^{l})}}
\{
u_{\alpha}(\hat{\underline{\underline{\theta}}}^{l}) \}.  \nonumber
\end{equation}

($ii$) If
$\Phi_{l}^{(\hat{B},\hat{\underline{\underline{\theta}}}^{l})}=\emptyset$,
take   $\pi^{0}(\hat{\underline{\underline{\theta}}}^{l})=b^l_0$, meaning that no bandit offers a better
solution under an increase,  the {Z-UCB policy $\pi^0$} employs the LPB($b^l_0$) block.\\ 

%Update the vector of estimates $\hat{\underline{\underline{\theta}}}^{l+1}$ and the statistics $S_{\pi^0}(l)$ and $T_{\pi^0}^{\alpha}(S_{\pi^0}(l))$. Repeat step 2 for $l:=l+1$. 

\end{tcolorbox}

\noindent{\bf Remark 2}  With apologies for the notation we have used $b^l_0$ as the initial optimal solution of  LP$(\hat{\underline{\underline{\theta}}}^{l})$, for the  $l$ block,  
and $b^0_{\alpha}(\hat{\underline{\underline{\theta}}}^{l})$ as the inflated solutions of  $b^0_l$ , cf.  \eqref{vi} and  \eqref{phi}.  
%Note also that  the statistics $S_{\pi^0}(l-1)$ and $T_{\pi^0}^{\alpha}(S_{\pi^0}(l-1))$ are updated according the their definition in section \ref{sec:breg}, only at the beginning of each block. 

  In   words, the Z-UCB policy firstly employs an ISB block in order to have estimates for each bandit. Then, for $l=1,2,\ldots$, it employs only LPB blocks according to the following. Firstly, the Z-UCB policy 
finds the  initial optimal solution  $b^l_0$ (with basis matrix  $\hat{B}$) which is a solution based on the estimates up to this block. Secondly, it  
computes the indices 
$u_{\alpha}(\hat{\underline{\underline{\theta}}}^{l},s,t )$
(of Eq. (\ref{eq:zucb})) 
for all bandits $\alpha \in 
\Phi_{l}^{(\hat{B},\hat{\underline{\underline{\theta}}}^{l})},$
 by varying only $\underline{\theta}_{\alpha}^{'}.$ 
Thus, for any  $\alpha \in 
\Phi_{l}^{(\hat{B},\hat{\underline{\underline{\theta}}}^{l})},$ we have the 
corresponding   new uniquely optimal BFS:  
 $b^0_{\alpha}(\hat{\underline{\underline{\theta}}}^{l}).$
 Then, the  Z-UCB policy chooses to employ the LPB block which corresponds to the highest  $z$ value among  $b^0_{\alpha}(\hat{\underline{\underline{\theta}}}^{l}), $  $\alpha \in 
\Phi_{l}^{(\hat{B},\hat{\underline{\underline{\theta}}}^{l})}.$\\

The main result of this paper is that under the following conditions policy $\pi^0$ is asymptotically optimal in the class of f-UF policies.

To state condition C1 below we need the definition of the bandit unobservable quantities: $J_{\alpha}(\underline{\underline{\theta}},\epsilon)$, as follows.   For any $\underline{\underline{\theta}} \in \Theta$,
 $ \epsilon>0$, an optimal matrix $B$ under $\underline{\underline{\theta}}$,
 as in Lemma 1, we define:
  $\Theta_{\alpha}^{'}(\epsilon)=\{\underline{\theta}_{\alpha}^{'} %\in \Theta_{\alpha\
: \,
\mu_{\alpha}^*(\underline{\underline{\theta}})-\epsilon <
\mu_{\alpha}(\underline{\theta}_{\alpha}^{'}) \}$ and
\begin{equation}
J_{\alpha}(\underline{\underline{\theta}},\epsilon)=
\inf_{\underline{\theta}_{\alpha}^{'}\in \Theta_{\alpha}^{'}(\epsilon)}\{
I(\underline{\theta}_{\alpha},\underline{\theta}_{\alpha}^{'}):
z(\underline{\theta}_{\alpha}^{'})>z^*
(\underline{\underline{\theta}})-\epsilon \}.\nonumber
\end{equation}

From the definition of 
$J_{\alpha}(\hat{\underline{\underline{\theta}}}^{l},\epsilon)$,
where
$\alpha\in\Phi_{l}^{(\hat{B},\hat{\underline{\underline{\theta}}}^{l})}$,

\begin{equation}
J_{\alpha}(\hat{\underline{\underline{\theta}}}^{l},\epsilon)=
\inf_{\underline{\theta}_{\alpha}^{'}}\{
I(\hat{\underline{\theta}}_{\alpha}^{l},\underline{\theta}_{\alpha}^{'}):
z^{b_{\alpha}(\hat{\underline{\underline{\theta}}}^{l},\underline{\theta}_{\alpha}^{'})}>z^*
(\underline{\underline{\theta}})-\epsilon \}, \nonumber
\end{equation}
we have that
$u_{\alpha}(\hat{\underline{\underline{\theta}}}^{l})>z^*
(\underline{\underline{\theta}})-\epsilon$ if and only if
$J_{\alpha}(\hat{\underline{\underline{\theta}}}^{l},\epsilon)< {\log
S_\pi(l-1)}/{T_\pi^{\alpha}(S_\pi(l-1))}$.\\

\noindent{\bf (C1) }
$\forall \ \underline{\underline{\theta}} \in
\Theta, \  i \notin b$ for any $ b \in O^*(\underline{\underline{\theta}})$  such
that $ \Delta\Theta_{i}(\underline{\underline{\theta}})=\emptyset,$
if
$\mu_{i}^*(\underline{\underline{\theta}})-\epsilon <
\mu_{i}(\underline{\theta}_{i}^{'}), $ \ $\forall \
\epsilon>0$, for some  $\underline{\theta}_{i}^{'} \in \Theta_{i}$, the following relation holds: $$ \lim_{\epsilon\rightarrow 0}
J_{i}(\underline{\underline{\theta}},\epsilon)=\infty.$$

\noindent{\bf (C2) } $  \forall i,$  $\forall \ \underline{\theta}_{i} \in \Theta_{i},
$ $\forall \ \epsilon > 0,$
$$P_{\underline{\theta}_i}(|\hat{\underline{\theta}}_{i}^{t}-\underline{\theta}_i|>\epsilon)=o(1/t), \mbox{ as } t\rightarrow \infty .$$

\noindent{\bf (C3) } $\forall i,$ $\forall \ \underline{\theta}_{i}
\in \Theta_{i}, $ $\forall \ \epsilon > 0,$ as $ t\rightarrow \infty
$
$$P_{\underline{\underline{\theta}}} (
z^{b_{i}(\hat{\underline{\underline{\theta}}}^{j},\underline{\theta}_{i}^{'})} \leq
z^* (\underline{\underline{\theta}})-\epsilon,
 \text{ for some }
j\leq t)=o(1/t).$$

 Next, we state and  prove the main theorem of the paper.\\

\begin{theorem}\label{th:ub}
Under conditions (C1),(C2), and (C3), and  policy $\pi^0$, defined above,  the following holds.
\begin{equation}
\uplim_{n\rightarrow\infty}\frac{R_{\pi^0}(\underline{\underline{\theta}},n)}{\log
n}\leq M(\underline{\underline{\theta}}), \text{ for all }
\underline{\underline{\theta}}\in\Theta.\nonumber
\end{equation}
\end{theorem}

\proof{%Proof of Theorem \ref{th:ub}.}
From \eqref{as2}, to establish the above inequality one can prove the same bound for $\uplim_{n\rightarrow\infty}\widetilde{R}_{\pi}(\underline{\underline{\theta}},L_n)/\log
L_n$. Now, from \eqref{tblock} and \eqref{eq:regb} we have the following 
{\small 
\begin{equation}
\widetilde{R}_{\pi}(\underline{\underline{\theta}},L_n)=
E_{\underline{\underline{\theta}}}(\sum_{j=1}^{k}\sum_{b\in K:j\in b}m_{j}^{b}\,\widetilde{T}_\pi^b(L_n)
+\sum_{j=1}^{k}m_{j}^{0})z^{*}(\underline{\underline{\theta}})+\sum_{j=1}^{k}m_{j}^{0}-
E_{\underline{\underline{\theta}}}\sum_{j=1}^{k}\sum_{b\in
K: j \in b}\mu_{j}(\underline{\theta}_j)\, m_{j}^{b}\,\widetilde{T}_\pi^b(L_n) \
-\sum_{j=1}^{k} \mu_{j}(\underline{\theta}_j)m^0_{j},\nonumber
\end{equation}
}
which using the relations: $z^*(\underline{\underline{\theta}})=\sum_{i=1}^{L}c^0_{i} g^{B}_{i}+g^{B}_{L+1}$ and
$\phi^{B}_{i} (\underline{\underline{\theta}}) =\sum_{j=1}^{L} c^{i}_{j}
g^{B}_{j}+ g^{B}_{L+1} - \mu_{i}(\underline{\theta}_{i})$ and after some algebra can be rewritten as:
\begin{equation}
\widetilde{R}_{\pi}(\underline{\underline{\theta}},L_n)=
\sum_{j=1}^{k}\sum_{b\in K:j\in b}m_{j}^{b}\,\phi^{B}_{j} (\underline{\underline{\theta}})E_{\underline{\underline{\theta}}}\widetilde{T}_\pi^b(L_n)
+\sum_{i=1}^{L} [ E_{\underline{\underline{\theta}}} S_{\pi^0} (L_n) c^0_i- E_{\underline{\underline{\theta}}}C_{i,\pi^0}(S_{\pi}(L_n))] g_{i}^{B}.\nonumber
\end{equation}

Now, since $\phi^{B}_{j} (\underline{\underline{\theta}})=0$ for optimal $j$, to prove the inequality for the regret it is sufficient to show that for  policy $\pi^0$ the relations below hold.

\begin{equation}\label{thm1}
\uplim_{n\rightarrow
\infty}\frac{E_{\underline{\underline{\theta}}}\widetilde{T}_{\pi^0,2}^b(L_n)}{\log
L_n} \leq\frac{1}{m_{i}^{b}K_{i}(\underline{\underline{\theta}})},
\text{ for all } i\in D(\underline{\underline{\theta}}), i\in b, b\notin O^*(\underline{\underline{\theta}}), 
\end{equation}
\begin{equation}\label{thm2}
\uplim_{n\rightarrow
\infty}\frac{E_{\underline{\underline{\theta}}}\widetilde{T}_{\pi^0,2}^b(L_n)}{\log
L_n}=0, \text{ for all } i\notin
D(\underline{\underline{\theta}}),i\in b, b\notin
O^*(\underline{\underline{\theta}}), 
\end{equation}
\begin{equation}\label{thm3}
\uplim_{n\rightarrow
\infty}\frac{E_{\underline{\underline{\theta}}}\widetilde{T}_{\pi^0,1}^b(L_n)}{\log
L_n}=0, \text{ for all } b\notin
O^*(\underline{\underline{\theta}}),
\end{equation}
\begin{equation}\label{thm4}
\sum_{i=1}^{L} [ E_{\underline{\underline{\theta}}} S_{\pi^0} (L_n) c^0_i- E_{\underline{\underline{\theta}}}C_{i,\pi^0}(S_{\pi}(L_n))] g_{i}^{B}=o(\log
n),  
\end{equation}
where  $ B$   is an  optimal basis under $\underline{\underline{\theta}}, $ and 
$\widetilde{T}_{\pi^0,1}^b(L_n)$ and $\widetilde{T}_{\pi^0,2}^b(L_n)$ can be obtained by the fragmentation  of $\widetilde{T}_{\pi^0}^b(L_n)$ as:
\begin{eqnarray}
\widetilde{T}_{\pi^0}^b(L_n)&=&\sum_{t=2}^{L_n}1\{ \pi_{t}^{0}=b,
b\notin
O^*(\underline{\underline{\theta}}), b \in O^*(\hat{\underline{\underline{\theta}}}^{t}),
u_{i}(\hat{\underline{\underline{\theta}}}^{t},\underline{\theta}_{i}^{'})=u_{\alpha^*}(\hat{\underline{\underline{\theta}}}^{t}) \}\nonumber\\
&=&\sum_{t=2}^{L_n}1\{ \pi_{t}^{0}=b,
b\notin
O^*(\underline{\underline{\theta}}), b \in O^*(\hat{\underline{\underline{\theta}}}^{t}),
u_{i}(\hat{\underline{\underline{\theta}}}^{t},\underline{\theta}_{i}^{'})=u_{\alpha^*}(\hat{\underline{\underline{\theta}}}^{t}),
u_{i}(\hat{\underline{\underline{\theta}}}^{t},\underline{\theta}_{i}^{'})\leq
z^* (\underline{\underline{\theta}})-\epsilon \} \nonumber \\
&&+\sum_{t=2}^{L_n}1\{ \pi_{t}^{0}=b,
b\notin
O^*(\underline{\underline{\theta}}), b \in O^*(\hat{\underline{\underline{\theta}}}^{t}),
u_{i}(\hat{\underline{\underline{\theta}}}^{t},\underline{\theta}_{i}^{'})=u_{\alpha^*}(\hat{\underline{\underline{\theta}}}^{t}),
u_{i}(\hat{\underline{\underline{\theta}}}^{t},\underline{\theta}_{i}^{'})> z^* (\underline{\underline{\theta}})-\epsilon \} \nonumber\\
&=&\widetilde{T}_{\pi^0,1}^b(L_n)+\widetilde{T}_{\pi^0,2}^b(L_n).\nonumber
\end{eqnarray}

%\begin{eqnarray}
%\widetilde{T}_\pi^b(L_n)&=&\sum_{t=2}^{L_n}1\{ \pi_{t}^{0}= b,
%b\notin
%O^*(\underline{\underline{\theta}}), b\notin O^*(\hat{\underline{\underline{\theta}}}^{t}) \}
%+ \sum_{t=2}^{L_n}1\{ \pi_{t}^{0}=b, b\notin
%O^*(\underline{\underline{\theta}}), b\in O^*(\hat{\underline{\underline{\theta}}}^{t}) \}\nonumber \\
%&\leq&\sum_{t=2}^{L_n}1\{
%b\notin
%O^*(\underline{\underline{\theta}}), b\notin O^*(\hat{\underline{\underline{\theta}}}^{t}) \} + \sum_{t=2}^{L_n}1\{
%\pi_{t}^{0}=b, b\notin
%O^*(\underline{\underline{\theta}}), b \in O^*(\hat{\underline{\underline{\theta}}}^{t}) \},\nonumber\\
%&=&\widetilde{T}_{\pi^0,1}^b(L_n)+\widetilde{T}_{\pi^0,2}^b(L_n).\nonumber
%\end{eqnarray}

The proof of the inequalities \eqref{thm1}, \eqref{thm2} and \eqref{thm3} is given in Lemma 7 in the appendix.
  \eqref{thm4} follows from Lemma $2$ which shows that for the constraints that do not obtain the optimal solution the corresponding $g^B$ are equal to $0$ and that a block based policy for the optimal bandits uses the whole amount $c^0$ of the constraints that give the optimal solution.
\endproof

\noindent{\bf Remark 3} According to Remark $4b$ in \cc{bkmab96}
 condition (C2) is equivalent to C2$'$ below which is often easier to verify.

\noindent{\bf (C2$'$)}
  $\forall \ \delta>0,$  as  $t\rightarrow \infty$
$$
 \sum_{j=1}^{t-1} P_{\underline{\theta}_i}
(b\notin
O^*(\underline{\underline{\theta}}), b \in O^*(\hat{\underline{\underline{\theta}}}^{j}),J_{i}(\hat{\underline{\underline{\theta}}}^{j},\epsilon)\leq
J_{i}(\underline{\underline{\theta}},\epsilon)-\delta) = o(\log t) .
$$
\\
\\
\section{Applications}
\subsection{Normal Distributions with Unknown Means and Known Variances}

Assume the observations $X_{\alpha}^{j}$ from bandit
$\alpha$ are normally distributed with unknown means
$EX_{\alpha}^{j}=\theta_{\alpha}$ and known variances
$\sigma_{\alpha}^{2}$, i.e.,
$\underline{\theta}_{\alpha}=\theta_{\alpha}$, $\underline{\underline{\theta}}=\underline{\theta}$,
$\mu_{\alpha}(\underline{\theta}_{\alpha})=\theta_{\alpha}$, and
$\Theta_{\alpha}=(0,+\infty)$.
Given history $h_{l}$, define
\begin{equation}
\mu_{\alpha}(\hat{\theta}_{\alpha}^{l})=
\frac{\sum_{j=1}^{T_{\pi^0}^{\alpha}(S_{\pi^0}(l-1))}
X_{\alpha}^{j}}{T_{\pi^0}^{\alpha}(S_{\pi^0}(l-1))}.\nonumber
\end{equation}

Now from the definition of $\Theta_{\alpha}$, it follows that
$\Delta\Theta_{\alpha}(\underline{\theta})=(\theta_{\alpha}+\phi_{\alpha}^{B}(\underline{\theta}),
\infty)$ for any optimal matrix $B$ under
$\underline{\theta}$, therefore $D(\underline{\theta})=\{ \alpha : \alpha \notin b \text{ for any } b \in O^*(\underline{\theta})\}$, $\forall$ $\underline{\theta}\in\Theta$. Thus, we can see from
the structure of the sets $\Theta_{\alpha}$ and
$\Delta\Theta_{\alpha}(\underline{\theta})$ that condition (C1)
is satisfied and that $\Phi_{l}^{(\hat{B},\hat{\underline{\theta}}^{l})}\neq\emptyset$ (we do not have the case (ii) in Step 2 of our policy).

Also, we have:
$$I(\theta_{\alpha},\theta_{\alpha}^{'})
=\frac{(\theta_{\alpha}^{'}-\theta_{\alpha})^2}{2\sigma_{\alpha}^{2}}$$
$$K_{\alpha}(\underline{\theta})=
\frac{(\phi_{\alpha}^{B}(\underline{\theta}))^2}{2\sigma_{\alpha}^{2}}.$$
It is easy to see that  the Z-UCB indices of Eq. (\ref{eq:zucb}) simplify to:  
$$
u_{\alpha}(\hat{\underline{\theta}}^{l})=
z^{b_{\alpha}(\hat{\underline{\theta}}^{l},\theta_{\alpha}^{K_{\alpha}})},$$
 where
$$ \theta_{\alpha}^{K_{\alpha}}=\hat{\theta}_{\alpha}^{l}+
\sigma_{\alpha}\left(\frac{2\log
S_{\pi^0}(l-1)}{T_{\pi^0}^{\alpha}(S_{\pi^0}(l-1))}\right)^{1/2},$$ 
is the $\theta_{\alpha}$ which satisfies the maximum in the index $u_{\alpha}(\hat{\underline{\theta}}^{l})$, of Eq. (\ref{eq:zucb}).

Note that
$b_{\alpha}(\hat{\underline{\theta}}^{l},\theta_{\alpha}^{K_{\alpha}})=\{i_1,\ldots,\alpha,\ldots,i_j\}$; thus, in the first case we have  
$z^{b_{\alpha}(\hat{\underline{\theta}}^{l},\theta_{\alpha}^{K_{\alpha}})}=\hat{\theta}_{i_1}^{l} x_{i_1}+\ldots+
\theta_{\alpha}^{K_{\alpha}} x_{\alpha} + \ldots+ \hat{\theta}_{i_j}^{l} x_{i_j}$ and $z^{*}(\underline{\theta})=\theta_{i_1}x_{i_1}+\ldots+\theta_{\alpha}x_{\alpha}+\ldots+\theta_{i_j}
x_{i_j}$.
 Therefore, for
$b_{\alpha}(\hat{\underline{\theta}}^{l},\theta_{\alpha}^{K_{\alpha}})\in
O^*(\underline{\theta})$ and from the structure of
$z^{b_{\alpha}(\hat{\underline{\theta}}^{l},\theta_{\alpha}^{K_{\alpha}})}$
the index is a sum of normal distributions which is also  a
normal distribution, and from a well known tail inequality  of normal distribution
condition (C3) is satisfied.

According to Remark 3 the next sum of probabilities is equivalent to
condition (C2)
\begin{eqnarray}
&&\sum_{t=2}^{L_n} P_{\theta_i} (b\notin
O^*(\underline{\underline{\theta}}), b \in O^*(\hat{\underline{\underline{\theta}}}^{t}),J_{i}(\hat{\underline{\theta}}^{t},\epsilon)\leq
J_{i}(\underline{\theta},\epsilon)-\delta)\nonumber\\
&& =\sum_{t=2}^{L_n}P_{\theta_i} (
b\notin
O^*(\underline{\underline{\theta}}), b \in O^*(\hat{\underline{\underline{\theta}}}^{t}),
|\hat{\theta}_{i}^{t}-\theta_i|>\xi ),\xi>0,\nonumber
\end{eqnarray}
where the equality follows after some algebra because of the normal
distribution and the explicit form of $I(\hat{\theta}_{i}^{t},\theta_{i}^{'})$ in this case:
\begin{eqnarray}
&&J_{i}(\hat{\underline{\theta}}^{t},\epsilon)=
\inf_{\theta_{i}^{'}}\{ I(\hat{\theta}_{i}^{t},\theta_{i}^{'}):
z^{b_{i}(\hat{\underline{\theta}}^{t},\theta_{i}^{'})}>z^*
(\underline{\theta})-\epsilon \} \leq \nonumber\\
&&J_{i}(\underline{\theta},\epsilon)=\inf_{\theta_{i}^{'}}\{
I(\theta_{i},\theta_{i}^{'}):
z^{b_{i}(\underline{\theta},\theta_{i}^{'})}>z^*
(\underline{\theta})-\epsilon \}-\delta.\nonumber
\end{eqnarray}
Also, we have that $\hat{\theta}_{i}^{t}$ is the average of iid
random normal variables with mean $\theta_i$ thus
\begin{eqnarray}
P_{\theta_i}^{\pi^0}(|\hat{\theta}_{i}^{t}-\theta_i|>\xi)& \leq&
P_{\theta_i}^{\pi^0}(|\hat{\theta}_{i}^{l}-\theta_i|>\xi,
\text{ for some }l\leq t)\nonumber\\
&\leq&\sum_{l=2}^{t}P_{\theta_i}^{\pi^0}(|\hat{\theta}_{i}^{l}-\theta_i|>\xi)=o(1/t),\nonumber
\end{eqnarray}
where the last equality follows from a consequence of the tail
inequality $1 - \Phi(x) < \Phi(x)/x$ for the standard normal
distribution, \cc{feller}. Thus, we can see that condition (C2) holds.\\

\noindent
\begin{tcolorbox}[colback=blue!1, arc=3pt, width=.94\linewidth]
{\bf Summary of Z-UCB Policy $\pi^0$:} 

{\bf Step 1}
Employ one ISB block  in 
order to have   one estimate $\hat{\theta}_a$ from each bandit $a$. Then, update the vector of estimates 
 $\hat{\underline{\theta}}^{2},$ and the statistics %\footnote[1]{$S_{\pi^0}(l)$ is the total length of first $l$ blocks (including the ISB block) and $T_{\pi^0}^{\alpha}(S_{\pi^0}(l))$ is the number of times bandit $\alpha$
%has been activated during the first $S_{\pi^0}(l)$ periods. Update them according to their definition in section \ref{sec:breg}.} 
%
$S_{\pi^0}(1)$ and $T_{\pi^0}^{\alpha}(S_{\pi^0}(1))$.\\
 
 {\bf Step 2} For block $l$ $(l>1)$ employ  an LPB($\pi^{0}(\hat{\underline{\theta}}^{l})$) block defined below. %according to 
 %a LPB($b^l$)  which corresponds to a feasible (not necessarily optimal solution $b^l=b^l(\hat{\underline{\underline{\theta}}}^{l})$ which is equal to 

 Given the history until the beginning of block $l$ we have the vector of estimates $\hat{\underline{\theta}}^{l}$ and the statistics $S_{\pi^0}(l-1)$ and $T_{\pi^0}^{\alpha}(S_{\pi^0}(l-1))$. Based on these, compute %\footnote[2]{The optimal policy of LP$(\hat{\underline{\underline{\theta}}}^{l})$ , the corresponding value of the objective function respectively.}
 $b^l_0$, and $z^{b^l_0}(\hat{\underline{\theta}}^{l})$. 
 Then for every bandit $\alpha=\{1,...,k\}$ we compute the indices
$$ u_{\alpha}(\hat{\underline{\theta}}^{l})=
z^{b_{\alpha}(\hat{\underline{\theta}}^{l},\theta_{\alpha}^{K_{\alpha}})},$$
where $b_{\alpha}(\hat{\underline{\theta}}^{l},\theta_{\alpha}^{K_{\alpha}})$ is the solution which we obtain if we replace only the parameter of bandit $\alpha$ by $\theta_{\alpha}^{K_{\alpha}}$, where
\begin{equation}\label{thetaka} \theta_{\alpha}^{K_{\alpha}}=\hat{\theta}_{\alpha}^{l}+
\sigma_{\alpha}\left(\frac{2\log
S_{\pi^0}(l-1)}{T_{\pi^0}^{\alpha}(S_{\pi^0}(l-1))}\right)^{1/2}.\nonumber
\end{equation}
The Z-UCB policy employs as block $l$ the 
$\pi^{0}(\hat{\underline{\theta}}^{l})=\argmax_{b_{\alpha}} \{
u_{\alpha}(\hat{\underline{\theta}}^{l})\}$, where ties in the $\argmax$ are broken arbitrarily. 

%Update the vector of estimates $\hat{\underline{\theta}}^{l+1}$ and the statistics $S_{\pi^0}(l)$ and $T_{\pi^0}^{\alpha}(S_{\pi^0}(l))$. Repeat step 2 for $l:=l+1$.
\end{tcolorbox}

\subsection{Normal Distributions with Unknown Means and Unknown Variances}

Assume the observations $X_{\alpha}^{j}$ from bandit
$\alpha$ are normally distributed with unknown means
$EX_{\alpha}^{j}=\mu_{\alpha}$ and unknown variances
$VarX_{\alpha}^{j}=\sigma^2_{\alpha}$, i.e.,
$\underline{\theta}_{\alpha}=(\mu_{\alpha},\sigma^2_{\alpha})$, and
$\Theta_{\alpha}=\{ \underline{\theta}_{\alpha} : \mu_{\alpha}>0, \ \sigma^2_{\alpha}\geq0\}$.
Given history $h_{l}$, define
\begin{equation}
\mu_{\alpha}(\hat{\underline{\theta}}_{\alpha}^{l})=
\frac{\sum_{j=1}^{T_{\pi^0}^{\alpha}(S_{\pi^0}(l-1))}
X_{\alpha}^{j}}{T_{\pi^0}^{\alpha}(S_{\pi^0}(l-1))} \mbox{ and } \sigma^2_{\alpha}(\hat{\underline{\theta}}_{\alpha}^{l})=
\frac{\sum_{j=1}^{T_{\pi^0}^{\alpha}(S_{\pi^0}(l-1))}
(X_{\alpha}^{j}-\mu_{\alpha}(\hat{\underline{\theta}}_{\alpha}^{l}))^2}{T_{\pi^0}^{\alpha}(S_{\pi^0}(l-1))}.\nonumber
\end{equation}

Now from the definition of $\Theta_{\alpha}$, it follows that
$\Delta\Theta_{\alpha}(\underline{\underline{\theta}})=(\mu_{\alpha}+\phi_{\alpha}^{B}(\underline{\underline{\theta}}),
\infty)$ for any optimal matrix $B$ under
$\underline{\underline{\theta}}$, therefore $D(\underline{\underline{\theta}})=\{ \alpha : \alpha \notin b \text{ for any } b \in O^*(\underline{\underline{\theta}})\}$, $\forall$ $\underline{\underline{\theta}}\in\Theta$. Thus, we can see from
the structure of the sets $\Theta_{\alpha}$ and
$\Delta\Theta_{\alpha}(\underline{\underline{\theta}})$ that condition (C1)
is satisfied and that $\Phi_{l}^{(\hat{B},\hat{\underline{\underline{\theta}}}^{l})}\neq\emptyset$ (we do not have the case (ii) in Step 2 of our policy).

Also, we have:
$$I(\underline{\theta}_{\alpha},\underline{\theta}_{\alpha}^{'})
=\frac{1}{2}\log\left(1+\frac{(\mu(\underline{\theta}_{\alpha}^{'})-\mu_{\alpha})^2}{\sigma_{\alpha}^{2}}\right)$$
$$K_{\alpha}(\underline{\underline{\theta}})=
\frac{1}{2}\log\left(1+\frac{(\phi_{\alpha}^{B}(\underline{\underline{\theta}}))^2}{\sigma_{\alpha}^{2}}\right).$$
It is easy to see that  the Z-UCB indices of Eq. (\ref{eq:zucb}) simplify to:  
$$
u_{\alpha}(\hat{\underline{\underline{\theta}}}^{l})=
z^{b_{\alpha}(\hat{\underline{\underline{\theta}}}^{l},\underline{\theta}_{\alpha}^{K_{\alpha}})},$$
 where
$$ \mu_{\alpha}(\underline{\theta}_{\alpha}^{K_{\alpha}})=\mu_{\alpha}(\hat{\underline{\theta}}_{\alpha}^{l})+
\sigma_{\alpha}(\hat{\underline{\theta}}_{\alpha}^{l})\left(S_{\pi^0}(l-1)^{\frac{2
}{T_{\pi^0}^{\alpha}(S_{\pi^0}(l-1))-2}}-1\right)^{1/2},$$ 
is the mean of the $\underline{\theta}_{\alpha}$ which satisfies the maximum in the index $u_{\alpha}(\hat{\underline{\underline{\theta}}}^{l})$, of Eq. (\ref{eq:zucb}).

Note that
$b_{\alpha}(\hat{\underline{\underline{\theta}}}^{l},\underline{\theta}_{\alpha}^{K_{\alpha}})=\{i_1,\ldots,\alpha,\ldots,i_j\}$; thus, in the first case we have  
$z^{b_{\alpha}(\hat{\underline{\underline{\theta}}}^{l},\underline{\theta}_{\alpha}^{K_{\alpha}})}=\mu_{i_1}(\hat{\underline{\theta}}_{i_1}^{l}) x_{i_1}+\ldots+
\mu_{\alpha}(\underline{\theta}_{\alpha}^{K_{\alpha}}) x_{\alpha} + \ldots+ \mu_{i_j}(\hat{\underline{\theta}}_{i_j}^{l})  x_{i_j}$ and $z^{*}(\underline{\underline{\theta}})=\mu_{i_1}x_{i_1}+\ldots+\mu_{\alpha}x_{\alpha}+\ldots+\mu_{i_j}
x_{i_j}$.
 Therefore, for
$b_{\alpha}(\hat{\underline{\underline{\theta}}}^{l},\underline{\theta}_{\alpha}^{K_{\alpha}})\in
O^*(\underline{\underline{\theta}})$ and from the structure of
$z^{b_{\alpha}(\hat{\underline{\underline{\theta}}}^{l},\underline{\theta}_{\alpha}^{K_{\alpha}})}$
the index is a sum of normal distributions which is also  a
normal distribution, and from a well known tail inequality  of normal distribution
condition (C3) is satisfied. Finally, condition (C2) is satisfied according to the analysis in \cite{chk2018}.\\

\noindent
\begin{tcolorbox}[colback=blue!1, arc=3pt, width=.94\linewidth]
{\bf Summary of Z-UCB Policy $\pi^0$:} 

{\bf Step 1}
Employ one ISB block  in 
order to have   one estimate $\hat{\underline{\theta}}_a$ from each bandit $a$. Then, update the vector of estimates 
 $\hat{\underline{\underline{\theta}}}^{2},$ and the statistics %\footnote[1]{$S_{\pi^0}(l)$ is the total length of first $l$ blocks (including the ISB block) and $T_{\pi^0}^{\alpha}(S_{\pi^0}(l))$ is the number of times bandit $\alpha$
%has been activated during the first $S_{\pi^0}(l)$ periods. Update them according to their definition in section \ref{sec:breg}.} 
%
$S_{\pi^0}(1)$ and $T_{\pi^0}^{\alpha}(S_{\pi^0}(1))$.\\
 
 {\bf Step 2} For block $l$ $(l>1)$ employ  an LPB($\pi^{0}(\hat{\underline{\underline{\theta}}}^{l})$) block defined below. %according to 
 %a LPB($b^l$)  which corresponds to a feasible (not necessarily optimal solution $b^l=b^l(\hat{\underline{\underline{\theta}}}^{l})$ which is equal to 

 Given the history until the beginning of block $l$ we have the vector of estimates $\hat{\underline{\underline{\theta}}}^{l}$ and the statistics $S_{\pi^0}(l-1)$ and $T_{\pi^0}^{\alpha}(S_{\pi^0}(l-1))$. Based on these, compute %\footnote[2]{The optimal policy of LP$(\hat{\underline{\underline{\theta}}}^{l})$ , the corresponding value of the objective function respectively.}
 $b^l_0$, and $z^{b^l_0}(\hat{\underline{\underline{\theta}}}^{l})$. 
 Then for every bandit $\alpha=\{1,...,k\}$ we compute the indices
$$ u_{\alpha}(\hat{\underline{\underline{\theta}}}^{l})=
z^{b_{\alpha}(\hat{\underline{\underline{\theta}}}^{l},\underline{\theta}_{\alpha}^{K_{\alpha}})},$$
where $b_{\alpha}(\hat{\underline{\underline{\theta}}}^{l},\underline{\theta}_{\alpha}^{K_{\alpha}})$ is the solution which we obtain if we replace only the parameter of bandit $\alpha$ by $\underline{\theta}_{\alpha}^{K_{\alpha}}$, where
\begin{equation}\label{thetaka1} \mu_{\alpha}(\underline{\theta}_{\alpha}^{K_{\alpha}})=\mu_{\alpha}(\hat{\underline{\theta}}_{\alpha}^{l})+
\sigma_{\alpha}(\hat{\underline{\theta}}_{\alpha}^{l})\left(S_{\pi^0}(l-1)^{\frac{2
}{T_{\pi^0}^{\alpha}(S_{\pi^0}(l-1))-2}}-1\right)^{1/2}.\nonumber
\end{equation}
The Z-UCB policy employs as block $l$ the 
$\pi^{0}(\hat{\underline{\underline{\theta}}}^{l})=\argmax_{b_{\alpha}} \{
u_{\alpha}(\hat{\underline{\underline{\theta}}}^{l})\}$, where ties in the $\argmax$ are broken arbitrarily. 

%Update the vector of estimates $\hat{\underline{\underline{\theta}}}^{l+1}$ and the statistics $S_{\pi^0}(l)$ and $T_{\pi^0}^{\alpha}(S_{\pi^0}(l))$. Repeat step 2 for $l:=l+1$.
\end{tcolorbox}

\subsection{Discrete Distributions with Finite Support}

Assume the observations $X_{\alpha}^{j}$ from bandit
$\alpha$ are univariate discrete distributions, i.e., $f_{\alpha}(x,\underline{p}_{\alpha})=p_{\alpha\, x}{\bf1} \{X_{\alpha}=x \}$, $x\in S_{\alpha}=\{ r_{\alpha 1},\ldots, r_{\alpha \,d_{\alpha}} \},$ where the parameters $p_{\alpha\,x}$ are unknown. The unknown parameters are in $\Theta_{\alpha}=\{ \underline{p}_{\alpha} \in {\Bbb R}^{d_{\alpha}}\,:\, p_{\alpha\,x}>0, \, \forall \, x=1,\ldots,d_{\alpha}, \, \sum_{x}p_{\alpha\, x}=1 \},$ and $r_{\alpha\, x}$ are known. Therefore, according to our notation, $\underline{\theta}_{\alpha}=\underline{p}_{\alpha}$ and $\underline{\underline{\theta}}=\underline{\underline{p}}=(\underline{p}_{1},\ldots,\underline{p}_{k})$. 

We can compute the mean reward of a bandit as $\mu_{\alpha}(\underline{\theta}_{\alpha})=\mu_{\alpha}(\underline{p}_{\alpha})=\underline{r}_{\alpha}^{'}\underline{p}_{\alpha}=\sum_{x}r_{\alpha\,x}p_{\alpha\,x}$, where $\underline{r}_{\alpha}^{'}$ denotes the transpose of the vector $\underline{r}_{\alpha}$. Now from the definition of $\Theta_{\alpha}$, it follows that
$\Delta\Theta_{\alpha}(\underline{\underline{p}})=(\mu_{\alpha}(\underline{p}_{\alpha})+\phi_{\alpha}^{B}(\underline{\underline{p}}),
\infty)$ for any optimal matrix $B$ under
$\underline{\underline{p}}$, therefore $D(\underline{\underline{p}})=\{ \alpha : \alpha \notin b \text{ for any } b \in O^*(\underline{\underline{p}})\}$, $\forall$ $\underline{\underline{p}}\in\Theta$. Thus, we can see from
the structure of the sets $\Theta_{\alpha}$ and
$\Delta\Theta_{\alpha}(\underline{\underline{p}})$ that condition (C1)
is satisfied and that $\Phi_{l}^{(\hat{B},\hat{\underline{\underline{p}}}^{l})}\neq\emptyset$ (we do not have the case (ii) in Step 2 of our policy).

Also, we can compute
$$
I(\underline{p}_{\alpha},\underline{p}_{\alpha}^{'})
=\sum_{x=1}^{d_{\alpha}}p_{\alpha\, x}\log \left(\frac{p_{\alpha\, x}}{p_{\alpha\, x}^{'}}\right),
$$
$$
K_{\alpha}(\underline{\underline{p}})=\min_{\underline{p}_{\alpha}^{'}}\{  I(\underline{p}_{\alpha},\underline{p}_{\alpha}^{'}) \, : \, \mu_{\alpha}(\underline{p}_{\alpha}^{'})\geq \mu_{\alpha}(\underline{p}_{\alpha})+\phi_{\alpha}^{B}(\underline{\underline{p}}), \, \sum_{x=1}^{d_{\alpha}} p_{\alpha\, x}^{'}=1\}.
$$

Now, for any estimators $\hat{\underline{p}}_{\alpha}^{l}$ of $\underline{p}_{\alpha}$, the computation of the index 
$$
u_{\alpha}(\hat{\underline{\underline{p}}}^{l})=u_{\alpha}(\hat{\underline{\underline{p}}}^{l},s,t )=
\max_{\underline{p}_{\alpha}^{'}}\{
z^{b_{\alpha}(\hat{\underline{\underline{p}}}^{l},\underline{p}_{\alpha}^{'})}:
I(\hat{\underline{p}}_{\alpha}^{l},\underline{p}_{\alpha}^{'})\leq\frac{\log
s}{t} \}, 
$$
involves the solution of $K_{\alpha}(\hat{\underline{\underline{p}}}^{l})$, which is a linear function maximization problem
subject to a constraint with convex level sets and a linear constraint, as in  \cc{bkmab96}. Thus, if the minimum in $K_{\alpha}(\hat{\underline{\underline{p}}}^{l})$ is $\underline{p}_{\alpha}^{K_{\alpha}}$ then
$$
u_{\alpha}(\hat{\underline{\underline{p}}}^{l})=
z^{b_{\alpha}(\hat{\underline{\underline{p}}}^{l},\underline{p}_{\alpha}^{K_{\alpha}})}. 
$$
Note that
$b_{\alpha}(\hat{\underline{\underline{p}}}^{l},\underline{p}_{\alpha}^{K_{\alpha}})=\{i_1,\ldots,\alpha,\ldots,i_j\}$; thus, in the first case we have  
$z^{b_{\alpha}(\hat{\underline{\underline{p}}}^{l},\underline{p}_{\alpha}^{K_{\alpha}})}=\underline{r}_{i_1}^{'}\hat{\underline{p}}_{i_1}^{l} x_{i_1}+\ldots+\underline{r}_{\alpha}^{'}
\underline{p}_{\alpha}^{K_{\alpha}} x_{\alpha} + \ldots+ \underline{r}_{i_j}^{'}\hat{\underline{p}}_{i_j}^{l} x_{i_j}$ and $z^{*}(\underline{\underline{p}})=\underline{r}_{i_1}^{'}\underline{p}_{i_1}x_{i_1}+\ldots+\underline{r}_{\alpha}^{'}\underline{p}_{\alpha}x_{\alpha}+\ldots+\underline{r}_{i_j}^{'}\underline{p}_{i_j}
x_{i_j}$. Thus, the index is just  a weighted sum of the estimated means and the inflated one, so in order to prove that the policy satisfies the conditions (C2) and (C3) one can follow exactly the same arguments that are used in Proposition 3 in \cc{bkmab96}. In Proposition 3 they use arguments based on the properties of the mean rewards of each bandit which hold in our case due to the form of our indices, as we analyzed above.

\noindent
\begin{tcolorbox}[colback=blue!1, arc=3pt, width=.94\linewidth]
{\bf Summary of Z-UCB Policy $\pi^0$:} 

{\bf Step 1}
Employ one ISB block  in 
order to have   one estimate $\hat{\underline{p}}_a$ from each bandit $a$. Then, update the vector of estimates 
 $\hat{\underline{\underline{p}}}^{2}$, and the statistics %\footnote[1]{$S_{\pi^0}(l)$ is the total length of first $l$ blocks (including the ISB block) and $T_{\pi^0}^{\alpha}(S_{\pi^0}(l))$ is the number of times bandit $\alpha$
%has been activated during the first $S_{\pi^0}(l)$ periods. Update them according to their definition in section \ref{sec:breg}.} 
%
$S_{\pi^0}(1)$ and $T_{\pi^0}^{\alpha}(S_{\pi^0}(1))$.\\
 
 {\bf Step 2} For block $l$ $(l>1)$ employ  an LPB($\pi^{0}(\hat{\underline{\underline{p}}}^{l})$) block defined below. %according to 
 %a LPB($b^l$)  which corresponds to a feasible (not necessarily optimal solution $b^l=b^l(\hat{\underline{\underline{\theta}}}^{l})$ which is equal to 

 Given the history until the beginning of block $l$ we have the vector of estimates $\hat{\underline{\underline{p}}}^{l}$ and the statistics $S_{\pi^0}(l-1)$ and $T_{\pi^0}^{\alpha}(S_{\pi^0}(l-1))$. Based on these, compute %\footnote[2]{The optimal policy of LP$(\hat{\underline{\underline{\theta}}}^{l})$ , the corresponding value of the objective function respectively.}
 $b^l_0$, and $z^{b^l_0}(\hat{\underline{\underline{p}}}^{l})$.

Then for every bandit $\alpha=\{1,...,k\}$ we compute the indices
$$ u_{\alpha}(\hat{\underline{\underline{p}}}^{l})=
z^{b_{\alpha}(\hat{\underline{\underline{p}}}^{l},\underline{p}_{\alpha}^{K_{\alpha}})} ,$$
where $b_{\alpha}(\hat{\underline{\underline{p}}}^{l},\underline{p}_{\alpha}^{K_{\alpha}})$ is the solution which we obtain if we replace only the parameter of bandit $\alpha$ by $\underline{p}_{\alpha}^{K_{\alpha}}$, where
\begin{equation}\label{pka} 
\underline{p}_{\alpha}^{K_{\alpha}}=\min_{\underline{p}_{\alpha}^{'}}\{  \sum_{x=1}^{d_{\alpha}}p_{\alpha\, x}\log \left(\frac{p_{\alpha\, x}}{p_{\alpha\, x}^{'}}\right) \, : \, \underline{r}_{\alpha}^{'}\underline{p}_{\alpha}^{'}= \underline{r}_{\alpha}^{'}\underline{p}_{\alpha}+\phi_{\alpha}^{B}(\underline{\underline{p}}), \, \sum_{x=1}^{d_{\alpha}} p_{\alpha\, x}^{'}=1\}.\nonumber
\end{equation}
The Z-UCB policy employs as block $l$   the  
$\pi^{0}(\hat{\underline{\underline{p}}}^{l})=\argmax_{b_{\alpha}} \{
u_{\alpha}(\hat{\underline{\underline{p}}}^{l})\}$,  where ties in the $\argmax$ are broken arbitrarily.

%Update the vector of estimates $\hat{\underline{\underline{p}}}^{l+1}$ and the statistics $S_{\pi^0}(l)$ and $T_{\pi^0}^{\alpha}(S_{\pi^0}(l))$. Repeat step 2 for $l:=l+1$. 
\end{tcolorbox}

\section{Conclusions}

This paper introduces a novel approach to multi-armed bandit (MAB) problems with side constraints, tackling the challenge of resource-limited activations. By deriving an asymptotic lower bound for regret in feasible policies and designing optimal strategies that achieve this bound, the study advances constrained bandit models. The proposed block-based Upper Confidence Bound (UCB) policies ensure asymptotic optimality while offering computational advantages over existing methods. These findings have broad implications for sequential decision-making in dynamic resource allocation, including applications in online revenue management and targeted advertising. Future research could refine these policies further, particularly in complex environments with evolving constraints and adaptive learning mechanisms.

\section{Compliance with Ethical Standards}
Conflict of Interest: The authors declare that there is no conflict of interests.

% Appendix here
% Options are (1) APPENDIX (with or without general title) or
%             (2) APPENDICES (if it has more than one unrelated sections)
% Outcomment the appropriate case if necessary
%
% 
\begin{appendix}{ }
\section{Appendix}
% {\bf Lemma 1} For any  optimal matrix $B$ under $\underline{\underline{\theta}}$, such
%that for any
%$\underline{\theta}_{\alpha}^{'}\in\Delta\Theta_{\alpha}(\underline{\underline{\theta}})$ the following is true,\\
%\ \ \  $\phi^B_{j}
%(\underline{\underline{\theta}}^{'})=\phi^B_{j}
%(\underline{\underline{\theta}})\geq 0, \ \forall \ j\neq\alpha $
%and $\phi^B_{\alpha}
%(\underline{\underline{\theta}}^{'})=\phi^B_{\alpha}
%(\underline{\underline{\theta}})
%+\mu_{\alpha}(\underline{\theta}_{\alpha})-\mu_{\alpha}(\underline{\theta}_{\alpha}^{'})<0$.

\proof{Lemma 1.}  It is obvious that $\phi^B_{j}
(\underline{\underline{\theta}}^{'})=\phi^B_{j}
(\underline{\underline{\theta}})\geq 0, \ \forall \ j\neq\alpha$
because we only change the parameter of bandit $\alpha$ and
$\phi^B_{j}
(\underline{\underline{\theta}}^{'})=\phi^B_{j}(\underline{\underline{\theta}})\equiv c^{j}_{1}
g^B_{1}+\ldots+c^{j}_{L}
g^B_{L}  + g^B_{L+1} - \mu_{j}(\underline{\theta}_{j})$.

For a bandit $\alpha\in B(\underline{\underline{\theta}})$ we
have that $\alpha\notin b$, for any $b\in
O^*(\underline{\underline{\theta}})$. Therefore $\phi^B_{\alpha}
(\underline{\underline{\theta}})\equiv c^{\alpha}_{1}
g^B_{1}+\ldots+c^{\alpha}_{L}
g^B_{L}  + g^B_{L+1} - \mu_{\alpha}(\underline{\theta}_{\alpha})
> 0$, for any $B$ corresponding to $b$.

Now, any optimal $b\in O^*(\underline{\underline{\theta}})$ is not
optimal under
$\underline{\underline{\theta}}^{'}=(\underline{\theta}_1,
\ldots,\underline{\theta}_{\alpha}^{'},\ldots,\underline{\theta}_k)$,
for any $\underline{\theta}_{\alpha}^{'}\in
\Delta\Theta_{\alpha}(\underline{\underline{\theta}})$, thus
$O^*(\underline{\underline{\theta}}^{'})=\{ b^{'} \}$ where
$b^{'}\notin O^*(\underline{\underline{\theta}})$.

Therefore, for any optimal matrix $B$ under
$\underline{\underline{\theta}}$ we have that $\phi^B_{\alpha}
(\underline{\underline{\theta}}^{'})\equiv  c^{\alpha}_{1}
g^B_{1}+\ldots+c^{\alpha}_{L}
g^B_{L}  + g^B_{L+1}  - \mu_{\alpha}(\underline{\theta}_{\alpha}^{'})< 0$ because $B$
is not optimal under $\underline{\underline{\theta}}^{'}$.

Now from $\phi^B_{\alpha} (\underline{\underline{\theta}})=
c^{\alpha}_{1}
g^B_{1}+\ldots+c^{\alpha}_{L}
g^B_{L}  + g^B_{L+1} -
\mu_{\alpha}(\underline{\theta}_{\alpha})$ we have that
$\phi^B_{\alpha}
(\underline{\underline{\theta}}^{'})=\phi^B_{\alpha}
(\underline{\underline{\theta}})
+\mu_{\alpha}(\underline{\theta}_{\alpha})-\mu_{\alpha}(\underline{\theta}_{\alpha}^{'})
< 0$.%$\blacksquare$
\endproof

%{\bf Lemma 2}If there is a uniquely optimal   $b\in
%O^*(\underline{\underline{\theta}})$. Then the following hold.\\ 
%$(i)$ If
%$b=\{i_1,\ldots,i_j\}$, $j=2,\ldots,L+1$,  then  $g^B_i >0$, for $j-1$ out of the $L$ resource   constraints, and equal to $0$ for the remaining  resource constraints.  
%
%$(ii)$ When $b$ is a singleton, i.e.,  $b=\{i_1\}$,  then  $g^B_i =0$, for every resource constraint  $i=1,\ldots,L$.

\proof{Lemma 2.} $(i)$ Let
$\underline{\underline{\theta}}:$
$O^*(\underline{\underline{\theta}})=\{ b \}$, then $g^B_i >0$, only for $j-1$ out of $L$ type of constraints because if $g^B_i =0$ for some of them we must have
more
than one solutions in the primal, which cannot occur because $b$ is uniquely optimal.\\
$(ii)$ Let $\underline{\underline{\theta}}:$
$O^*(\underline{\underline{\theta}})=\{ b \}$,
then $g^B_i =0$, for all $i=1,\ldots,L$ from the dual solution and
$\phi_{j}^{B}(\underline{\underline{\theta}})>0$ for all $j\neq
i_1$. Thus, $g^B_{L+1} $ is positive since the means ($\mu$'s) are positive.%$\blacksquare$
\endproof

\proof{Proposition \ref{prop:lb}.} Let $\alpha\in D(\underline{\underline{\theta}})$,
$\theta_{\alpha}^{'}\in\Delta\Theta_{\alpha}(\underline{\underline{\theta}}).$
The definition of $\Delta\Theta_{\alpha}(\underline{\underline{\theta}})$ and Lemma \ref{lem:bd1} 
imply that  we must have a $b^{'}$ which is uniquely optimal under
$\underline{\underline{\theta}}^{'}$
(i.e., $O^*(\underline{\underline{\theta}}^{'})=\{ b^{'} \}$) and $\alpha\in b^{'}$. Then we have two cases for the uniquely optimal solution
 $b^{'}$  depending on wether $b^{'}$ is a singleton or not.

In the first case   $b^{'}=\{ \alpha \}$,  
  and   Lemma \ref{lem:g} implies that for  a  f-UF policy $g^{B'}_i=0$, $i=1,\ldots,L$ thus, the definition of a f-UF policy implies that:
\begin{equation}\label{eq:on}
E_{\underline{\underline{\theta}}^{'}}T^{j}_{\pi}(n)=o(n^{a}),
\text{ for all } a>0, \text{ for all } j\notin b^{'}. 
\end{equation}
%If $b^{'}=\{ \alpha,j \}$ is degenerate then it must be true that $c^{\alpha}_i=c^0_i$
%for only one type of constraint, then $g^{B^{'}}_i>0$ thus
%$(c^0_i-c^{j}_i)E_{\underline{\underline{\theta}}^{'}}T^{j}_{\pi}(n)+
%(c^0_i
%-c^{\alpha}_i)E_{\underline{\underline{\theta}}^{'}}T^{\alpha}_{\pi}(n)
%=o(n^{a})$ and since $c^0_i=c^{\alpha}_i$ we have that
%$E_{\underline{\underline{\theta}}^{'}}T^{j}_{\pi}(n)=o(n^{a})$ also
%from \eqref{ug1}
%$E_{\underline{\underline{\theta}}^{'}}T^{i}_{\pi}(n)=o(n^{a})$, for
%all $i\neq j,\alpha$ thus
%$E_{\underline{\underline{\theta}}^{'}}T^{j}_{\pi}(n)=o(n^{a})$, for
%all $j\neq \alpha$.

Therefore,
\begin{equation}
n-E_{\underline{\underline{\theta}}^{'}}T^{\alpha}_{\pi}(n)=
\sum _{j\notin \bp} E_{\underline{\underline{\theta}}^{'}}T^{j}_{\pi}(n)
=o(n^{a}), \text{ for all } a>0.\label{prop1}
\end{equation}

Now for any sequence $\beta_n =o(n)$ with $\beta_n<n$ (for all $n$), we obtain the following.
\begin{eqnarray}
E_{\underline{\underline{\theta}}^{'}}T^{\alpha}_{\pi}(n) &=&
\sum_{k=1}^{n}k\, P_{\underline{\underline{\theta}}^{'}}[T^{\alpha}_{\pi}(n) =k]\nonumber\\
&=&\sum_{k=1}^{\lfloor\beta_n\rfloor}k\,
P_{\underline{\underline{\theta}}^{'}}[T^{\alpha}_{\pi}(n)
=k]+\sum_{k=\lfloor\beta_n\rfloor+1}^{n}k\,
P_{\underline{\underline{\theta}}^{'}}[T^{\alpha}_{\pi}(n) =k] \nonumber\\
&\leq&\beta_n
P_{\underline{\underline{\theta}}^{'}}[T^{\alpha}_{\pi}(n) \leq
\beta_n]+
n P_{\underline{\underline{\theta}}^{'}}[T^{\alpha}_{\pi}(n) >\beta_n]\nonumber\\
&=&n-(n-\beta_n)P_{\underline{\underline{\theta}}^{'}}[T^{\alpha}_{\pi}(n)
\leq \beta_n].\nonumber
\end{eqnarray}
Therefore
\begin{equation}
n-E_{\underline{\underline{\theta}}^{'}}T^{\alpha}_{\pi}(n)\geq(n-\beta_n)P_{\underline{\underline{\theta}}^{'}}[T^{\alpha}_{\pi}(n)\leq
\beta_n].\label{prop2}
\end{equation}

From \eqref{prop1} and \eqref{prop2} we obtain
\begin{equation}
(n-\beta_n)P_{\underline{\underline{\theta}}^{'}}[T^{\alpha}_{\pi}(n)
\leq \beta_n]=o(n^a), \text{ for all }a>0, \nonumber
\end{equation}
thus
\begin{equation}
P_{\underline{\underline{\theta}}^{'}}[T^{\alpha}_{\pi}(n) \leq
\beta_n]=o(n^{a-1}), \text{ for all }a>0. \nonumber
\end{equation}
And the proof is complete for this case. 

In the second case  $b^{'}=\{ i_1,\ldots,i_j\}$, for $j=2,\ldots,L+1$, where bandit $\alpha$ is one of $ i_1,\ldots,i_j$.
Then as before \eqref{eq:on} holds $\forall \ a>0, \ \forall \ i\notin \bp . $
It follows  from Lemma \ref{lem:g} that for an f-UF 
policy we must have $g^{B^{'}}_{i}>0$, for $j-1$ resource  constraints, which we label as  $i=s_1,\ldots,s_{j-1}.$   Using the last result  and 
 and \eqref{ug2} we obtain:
\begin{equation}
\sum_{i=s_1}^{s_{j-1}}g^{B^{'}}_{i}\sum_{l\in \bp}(c^0_{i} -
c^l_{i})E_{\underline{\underline{\theta}}^{'}}T_{\pi}^{l}(n)
=o(n^{a}), \ \forall \ a>0.\label{lem2}
\end{equation}

If we sum \eqref{eq:on} for all $i\notin b^{'}$ it follows that
\begin{equation}
n-\sum_{i\in \bp}E_{\underline{\underline{\theta}}^{'}}T_{\pi}^{i}(n)
=\sum_{i\notin \bp}E_{\underline{\underline{\theta}}^{'}}T_{\pi}^{i}(n)
=\varepsilon_n, \text{ where } \varepsilon_n=o(n^{a}), \ \forall \
a>0.\label{lem3}
\end{equation}

Now, let  $x_{i_1},\ldots,x_{i_j}$ be the corresponding randomization probabilities, then $\sum_{k \in \bp} x_{k} =1$ and from \eqref{lem3} we have that
\begin{equation}
\sum_{i\in \bp}(nx_i-E_{\underline{\underline{\theta}}^{'}}T_{\pi}^{i}(n)
)=\varepsilon_n, \text{ where } \varepsilon_n=o(n^{a}), \ \forall \
a>0.\label{lem3a}
\end {equation} 

From the definition of $z^*$  we can write 
$z^*(\underline{\underline{\theta}}^{'})=\sum_{i_k\neq \alpha} x_{i_k}\mu_{i_k}(\underline{\theta}_{i_k})+x_{\alpha}\mu_{\alpha}(\underline{\theta}_{\alpha}^{'})$,
and from the DLP we have  $z^*(\underline{\underline{\theta}}^{'})=\sum_{k=s_1}^{s_{j-1}}c^0_{k} g^{B^{'}}_{k}+g^{B^{'}}_{L+1}$.
Also, from the DLP we  obtain that 
$\phi^{B^{'}}_{i} (\underline{\underline{\theta}}^{'}) =\sum_{k=s_1}^{s_{j-1}} c^{i}_{k}
g^{B^{'}}_{k}+ g^{B^{'}}_{L+1} - \mu_{i}(\underline{\theta}_{i})$, for $i\neq \alpha,$ and 
$\phi^{B^{'}}_{\alpha} (\underline{\underline{\theta}}^{'}) =\sum_{k=s_1}^{s_{j-1}} c^{i}_{k}
g^{B^{'}}_{k}+ g^{B^{'}}_{L+1} - \mu_{i}(\underline{\theta}_{\alpha}^{'})$, 
where $\phi^{B^{'}}_{i} (\underline{\underline{\theta}}^{'})=0$ for $i\in b^{'}$. 

After some algebra we can show that
\begin{equation}\label{lem3aa}
z^*(\underline{\underline{\theta}}^{'})=(c^0_{s_1}- c^{i}_{s_1})g^{B^{'}}_{s_1}+\ldots+(c^0_{s_j}-c^{i}_{s_{j-1}} )g^{B^{'}}_{s_{j-1}}+\mu_{i}(\underline{\theta}_{i}), \text{ for every } i\in b^{'},
\end{equation}
and
\begin{eqnarray}\label{eq:zs}
z^*(\underline{\underline{\theta}}^{'})&=&\sum_{i_k\neq \alpha} x_{i_k}\mu_{i_k}(\underline{\theta}_{i_k})+x_{\alpha}\mu_{\alpha}(\underline{\theta}_{\alpha}^{'})\nonumber \\
&=&x_{i_1}(c^{i_1}_{s_1} g^{B^{'}}_{s_1}+\ldots+c^{i_1}_{s_{j-1}} g^{B^{'}}_{s_{j-1}} +g^{B^{'}}_{L+1})+\ldots+x_{i_j}(c^{i_j}_{s_1} g^{B^{'}}_{s_1}+\ldots+c^{i_j}_{s_{j-1}} g^{B^{'}}_{s_{j-1}} +g^{B^{'}}_{L+1}) \nonumber \\
&=&\sum_{i=s_1}^{s_{j-1}}g^{B^{'}}_{i}\sum_{l\in \bp}
c^l_{i}x_{i_l}+g^{B^{'}}_{L+1} . 
\end{eqnarray}

Using  $z^*(\underline{\underline{\theta}}^{'})=\sum_{k=s_1}^{s_{j-1}}c^0_{k} g^{B^{'}}_{k}+g^{B^{'}}_{L+1}$ , and  \eqref{eq:zs} we 
have:
 $z^*(\underline{\underline{\theta}}^{'})-g^{B^{'}}_{L+1}=\sum_{k=s_1}^{s_{j-1}}c^0_{k} g^{B^{'}}_{k}$ and \\ 
 $z^*(\underline{\underline{\theta}}^{'})-g^{B^{'}}_{L+1}=\sum_{i=s_1}^{s_{j-1}}g^{B^{'}}_{i}\sum_{l\in \bp}
c^l_{i}x_{i_l}$ which imply:
\begin{equation}\label{eq:zo1}
\sum_{i=s_1}^{s_{j-1}}g^{B^{'}}_{i}
c^0_{i}-\sum_{i=s_1}^{s_{j-1}}g^{B^{'}}_{i}\sum_{l\in \bp}
c^l_{i}x_{i_l}=0. 
\end{equation}
In addition, since $\sum_{k \in \bp} x_{k} =1$   \eqref{eq:zo1} can be written as:
\begin{equation}
\sum_{i=s_1}^{s_{j-1}}g^{B^{'}}_{i}
\sum_{l\in \bp}
c^0_{i}x_{i_l}-\sum_{i=s_1}^{s_{j-1}}g^{B^{'}}_{i}\sum_{l\in \bp}
c^l_{i}x_{i_l}=0,\nonumber
\end{equation}
which simplifies into: 
\begin{equation}
\sum_{i=s_1}^{s_{j-1}}g^{B^{'}}_{i}\sum_{l\in \bp}(c^0_{i} -
c^l_{i})x_{i_l}
=0.\nonumber
\end{equation}

Multiplying both sides of  the last equation by $n$,
\begin{equation}
\sum_{i=s_1}^{s_{j-1}}g^{B^{'}}_{i}\sum_{l\in \bp}(c^0_{i} -
c^l_{i})nx_{i_l}
=0.\label{lem3b}
\end{equation}

From \eqref{lem2} and \eqref{lem3b} we have that
\begin{equation}
\sum_{i=s_1}^{s_{j-1}}g^{B^{'}}_{i}\sum_{l\in \bp}(c^0_{i} -
c^l_{i})(nx_l-E_{\underline{\underline{\theta}}^{'}}T_{\pi}^{l}(n))
=o(n^{a}), \ \forall \ a>0.\label{lem3c}
\end{equation}

%Thus we have that 
%\begin{equation}
%z^*(\underline{\underline{\theta}}^{'})=(c^0_{s_1}- c^{i}_{s_1})g^{B^{'}}_{s_1}+\ldots+(c^0_{s_j}-c^{i}_{s_{j-1}} )g^{B^{'}}_{s_{j-1}}+\mu_{i}(\underline{\theta}_{i}), \text{ for every } i\in b^{'}.\label{new1}
%\end{equation}

%From \eqref{lem2} and \eqref{new1} we have 
%\begin{equation}
%\sum_{l\in b}(z^*(\underline{\underline{\theta}}^{'})-\mu_{l}(\underline{\theta}_{l}))E_{\underline{\underline{\theta}}^{'}}T_{\pi}^{l}(n)=o(n^{a}),
%\ \forall \ a>0.\label{new2}
%\end{equation}

%Now, multiplying \eqref{lem3} by $z^*(\underline{\underline{\theta}}^{'})$ and from \eqref{new2} 
%\begin{equation}
%\sum_{l\in b}\mu_{l}(\underline{\theta}_{l})(nx_l-E_{\underline{\underline{\theta}}^{'}}T_{\pi}^{l}(n))=o(n^{a}),
%\ \forall \ a>0,\nonumber
%\end{equation}

Combining, \eqref{lem3a} and \eqref{lem3c} we obtain,
\begin{equation}
nx_i-E_{\underline{\underline{\theta}}^{'}}T_{\pi}^{i}(n)=o(n^{a}),
\ \forall \ a>0, \ \forall \ i\in b^{'}.\label{new3} 
\end{equation}

For any $n$ let
\begin{equation}
\Gamma_{n}^{\pi}=\sum_{ l \notin b^{'}}T^{l}_{\pi}(n), \text{and }
F_{n,i}^{\pi} =\sum_{l\notin b^{'}}(c^0_{i} -c^l_{i})T^{l}_{\pi}(n),\nonumber
\end{equation}
where $i$ is the $i-$th resource constraint.  
Now, for each resource constraint $i$,  label as $c_i^*$  the minimum $c_i^j$ for $j=1,\ldots, k$. With thus defined  $c_i^*$ and the above definitions we have:
\begin{equation}
F_{n,i}^{\pi}\leq\Gamma_{n}^{\pi}(c^0_{i}-c^*_{i}), \text{ for all constraints } i=1,\ldots,L. \nonumber
\end{equation}

Furthermore, from \eqref{lem3}
\begin{equation}
E_{\underline{\underline{\theta}}^{'}}\Gamma_{n}^{\pi}=o(n^{a}), \  
\forall \ a>0.\label{lem5}
\end{equation}

Now, \eqref{eq:ccon} and the definition of $F_{n,i}^{\pi}$ imply the following. 
\begin{equation}\label{eq:arg1}
\sum_{i=s_1}^{s_{j-1}}g^{B^{'}}_{i}(nc^0_{i}-C_{i,\pi}(n))=\sum_{i=s_1}^{s_{j-1}}g^{B^{'}}_{i}F_{n,i}^{\pi} 
+\sum_{i=s_1}^{s_{j-1}}g^{B^{'}}_{i}\sum_{l\in b^{'}}(c^0_{i} -
c^l_{i})T_{\pi}^{l}(n). 
\end{equation}
For any  bandit in $ b^{'}$ the following arguments hold. For simplicity we present these arguments only for the specific bandit $\alpha$ for which $\alpha\in D(\underline{\underline{\theta}})$,
$\theta_{\alpha}^{'}\in\Delta\Theta_{\alpha}(\underline{\underline{\theta}}).$ 

Since   $g^{B^{'}}_{i}>0$ (from the optimality of $B^{'}$)  and $nc^0_i-C_{i,\pi}(n)\geq0, \ \forall \ n$,$\forall \ i$ (from the feasibility  of $\pi$) the right side of 
 \eqref{eq:arg1} is nonnegative.     Using  the  non negativity inequality of  the right side of 
 \eqref{eq:arg1} and moving the term that corresponds to bandit $\alpha$ to the left side in the equation  \eqref{eq:arg1} (and changing the sign of $(c^0_{\alpha} -c^l_{\alpha})$)  we obtain the inequality below,
\begin{equation}
\sum_{i=s_1}^{s_{j-1}}g^{B^{'}}_{i}(c^{\alpha}_{i}-c^0_{i})T_{\pi}^{\alpha}(n) \leq \sum_{i=s_1}^{s_{j-1}}g^{B^{'}}_{i}F_{n,i}^{\pi} 
+\sum_{i=s_1}^{s_{j-1}}g^{B^{'}}_{i}\sum_{l\in \bp, l\neq \alpha}
(c^0_{i} -c^l_{i})T_{\pi}^{l}(n).\nonumber
\end{equation}

Using  \eqref{lem3aa} on both sides of the above inequality (for $i=\alpha$ in the left hand side and for all  $i\in \bp, $ $i\neq \alpha$ in the right side)
 we obtain:
\begin{equation}
(\mu_{\alpha}(\underline{\theta}_{\alpha}^{'})-z^*(\underline{\underline{\theta}}^{'}))T^{\alpha}_{\pi}(n)  \leq  \sum_{i=s_1}^{s_{j-1}}g^{B^{'}}_{i}F_{n,i}^{\pi} 
+\sum_{l\in b^{'}, l\neq \alpha}(z^*(\underline{\underline{\theta}}^{'})-\mu_{l}(\underline{\theta}_{l}))T_{\pi}^{l}(n) ,\nonumber
\end{equation}
which simplifies into:
\begin{equation}
\mu_{\alpha}(\underline{\theta}_{\alpha}^{'}) T^{\alpha}_{\pi}(n)  \leq  \sum_{i=s_1}^{s_{j-1}}g^{B^{'}}_{i}F_{n,i}^{\pi}+ z^*(\underline{\underline{\theta}}^{'})\sum_{l\in b^{'}}T_{\pi}^{l}(n) -\sum_{l\in b^{'}, l\neq \alpha}\mu_{l}(\underline{\theta}_{l})T_{\pi}^{l}(n) . \nonumber
\end{equation}
Now from the definition of $\Gamma_{n}^{\pi}$ we have that $n-\Gamma_{n}^{\pi}=\sum_{l\in b^{'}}T_{\pi}^{l}(n)$ and recall that  $z^*(\underline{\underline{\theta}}^{'})=x_{\alpha}\mu_{\alpha}(\underline{\theta}_{\alpha}^{'})+\sum_{l\in b^{'}, l\neq \alpha}x_l \mu_{l}(\underline{\theta}_{l})$ and that by assumption $\mu_{l}(\underline{\theta}_{l})>0$. Using these and  simple  algebra   the above inequality can be written as, 
\begin{equation}
T^{\alpha}_{\pi}(n) \leq n
x_{\alpha}+ \sum_{i=s_1}^{s_{j-1}}\frac{g^{B^{'}}_{i}}{\mu_{\alpha}(\underline{\theta}_{\alpha}^{'})}F_{n,i}^{\pi} +\frac{1}{\mu_{\alpha}(\underline{\theta}_{\alpha}^{'})}\sum_{l\in b^{'},l\neq \alpha}(nx_l-T_{\pi}^{l}(n))\mu_{l}(\underline{\theta}_{l}).\nonumber
\end{equation}

%Now, we recall $F_n^{\pi}\leq\Gamma_n^{\pi}(c^0-c^1)$, thus
%\begin{eqnarray}
%T^{\alpha}_{\pi}(n) &\leq&n x_{\alpha}^{'}+
%\frac{\Gamma_n^{\pi}(c^0-c^1)}{c^{\alpha}-c^{j_0}}- x_{\alpha}^{'}\Gamma_n\nonumber\\
%T^{\alpha}_{\pi}(n) &\leq&n x_{\alpha}^{'} +\Gamma_n^{\pi}
%\rho(j_0,\alpha)\nonumber
%\end{eqnarray}
%where $\rho(j_0,\alpha)=\frac{c^{j_0}-c^1}{c^{\alpha}-c^{j_0}}\geq
%0$.

The last inequality can be rearranged and written as,
\begin{equation}
nx_{\alpha}-T^{\alpha}_{\pi}(n) + \sum_{i=s_1}^{s_{j-1}}\frac{g^{B^{'}}_{i}}{\mu_{\alpha}(\underline{\theta}_{\alpha}^{'})}F_{n,i}^{\pi} +\frac{1}{\mu_{\alpha}(\underline{\theta}_{\alpha}^{'})}\sum_{l\in b^{'}, l\neq \alpha}(nx_l-T_{\pi}^{l}(n))\mu_{l}(\underline{\theta}_{l})\geq 0 . \nonumber
\end{equation}
For simplicity  the above will be written as 
\begin{equation}
nx_{\alpha}-T^{\alpha}_{\pi}(n) + A^1_\pi (n) + A^2_\pi (n) \geq 0. \nonumber
\end{equation}
where $ A^1_\pi (n) =\sum_{i=s_1}^{s_{j-1}}\frac{g^{B^{'}}_{i}}{\mu_{\alpha}(\underline{\theta}_{\alpha}^{'})}F_{n,i}^{\pi} $ and $ A^2_\pi (n) =\frac{1}{\mu_{\alpha}(\underline{\theta}_{\alpha}^{'})}\sum_{l\in b^{'}, l\neq \alpha}(nx_l-T_{\pi}^{l}(n))\mu_{l}(\underline{\theta}_{l})$.

Note that from \eqref{new3} (and \eqref{lem5} respectively)  it follows that  $ A^i_\pi (n) =o(n^{a}), \ \forall \ a>0,$ and $i=1,2,$ since all  means are  positive.
From the  Markov inequality, for any positive $\beta_n=o(n)$
\begin{eqnarray}
P_{\underline{\underline{\theta}}^{'}}(nx_{\alpha}-T^{\alpha}_{\pi}(n) +  A^1_\pi (n) + A^2_\pi (n) \geq n x_{\alpha}-\beta_n) 
& \leq &\frac{E_{\underline{\underline{\theta}}^{'}}(nx_{\alpha}-T^{\alpha}_{\pi}(n) +  A^1_\pi (n) + A^2_\pi (n) )}{n x_{\alpha}^{'}-\beta_n}\nonumber\\
&= & \frac{o(n^{a})}{n x_{\alpha}-\beta_n}=o(n^{a-1}), \ \forall \
a>0.\nonumber
\end{eqnarray}
Using the above we obtain,
\begin{equation}
P_{\underline{\underline{\theta}}^{'}}(T^{\alpha}_{\pi}(n) \leq
\beta_n)\leq
P_{\underline{\underline{\theta}}^{'}}(T^{\alpha}_{\pi}(n) \leq
\beta_n +  A^1_\pi (n) + A^2_\pi (n) ) = o(n^{a-1}), \forall \
a>0.\nonumber
\end{equation}
And the proof is complete for this case too. %$\blacksquare$
\endproof 
%%%%****

For the analysis in proof of Lemma 4 we use the index $u_{i}(\hat{\underline{\underline{\theta}}}^{t},\underline{\theta}_{i}^{'})$ at block $t$, where $\underline{\theta}_{i}^{'}$ is the corresponding inflated $\underline{\theta}_{i}$, to denote the index $u_{i}(\hat{\underline{\underline{\theta}}}^{t})$ of bandit $i$. In fact $\underline{\theta}_{i}^{'} \equiv \underline{\theta}_{i}^0(
S_{\pi^0}(t-1),T_{\pi^0}^{i}(S_{\pi^0}(t-1)))$ according to the definition of our policy.

\begin{lemma} 
Under conditions (C1),(C2), and (C3) policy $\pi^0$ satisfies:\\
\begin{equation}
\uplim_{n\rightarrow
\infty}\frac{E_{\underline{\underline{\theta}}}\widetilde{T}_{\pi^0,2}^b(L_n)}{\log
L_n} \leq\frac{1}{m_{i}^{b}K_{i}(\underline{\underline{\theta}})},
\text{ for all } i\in D(\underline{\underline{\theta}}), i\in b, b\notin O^*(\underline{\underline{\theta}}),\nonumber
\end{equation}
\begin{equation}
\uplim_{n\rightarrow
\infty}\frac{E_{\underline{\underline{\theta}}}\widetilde{T}_{\pi^0,2}^b(L_n)}{\log
L_n}=0, \text{ for all } i\notin
D(\underline{\underline{\theta}}),i\in b, b\notin
O^*(\underline{\underline{\theta}}),\nonumber
\end{equation}
\begin{equation}
\uplim_{n\rightarrow
\infty}\frac{E_{\underline{\underline{\theta}}}\widetilde{T}_{\pi^0,1}^b(L_n)}{\log
L_n}=0, \text{ for all } b\notin
O^*(\underline{\underline{\theta}}).\nonumber
\end{equation}
\end{lemma}

\proof{} From the relation between the two indices $u_i$ and $J_i$ we have
that
\begin{gather}
\widetilde{T}_{\pi^0,2}^b(L_n)=\sum_{t=2}^{L_n}1\{ \pi_{t}^{0}=b,
b\notin
O^*(\underline{\underline{\theta}}), b \in O^*(\hat{\underline{\underline{\theta}}}^{t}),
u_{i}(\hat{\underline{\underline{\theta}}}^{t},\underline{\theta}_{i}^{'})=u_{\alpha^*}(\hat{\underline{\underline{\theta}}}^{t}),
 u_{i}(\hat{\underline{\underline{\theta}}}^{t},\underline{\theta}_{i}^{'})> z^* (\underline{\underline{\theta}})-\epsilon \} \nonumber\\
\leq\sum_{t=2}^{L_n}1\{ \pi_{t}^{0}=b,
b\notin
O^*(\underline{\underline{\theta}}), b \in O^*(\hat{\underline{\underline{\theta}}}^{t}),
u_{i}(\hat{\underline{\underline{\theta}}}^{t},\underline{\theta}_{i}^{'})=u_{\alpha^*}(\hat{\underline{\underline{\theta}}}^{t}),
 J_{i}(\hat{\underline{\underline{\theta}}}^{t},\epsilon)<\frac{\log
S_{\pi^0}(t-1)}{T_{\pi^0}^{i}(S_{\pi^0}(t-1))} \} \nonumber\\
=\sum_{t=2}^{L_n}1\{ \pi_{t}^{0}=b,
b\notin
O^*(\underline{\underline{\theta}}), b \in O^*(\hat{\underline{\underline{\theta}}}^{t}),
u_{i}(\hat{\underline{\underline{\theta}}}^{t},\underline{\theta}_{i}^{'})=u_{\alpha^*}(\hat{\underline{\underline{\theta}}}^{t}) ,\nonumber\\
J_{i}(\hat{\underline{\underline{\theta}}}^{t},\epsilon)<\frac{\log
S_{\pi^0}(t-1)}{T_{\pi^0}^{i}(S_{\pi^0}(t-1))},
 J_{i}(\hat{\underline{\underline{\theta}}}^{t},\epsilon)>J_{i}(\underline{\underline{\theta}},\epsilon)-\delta \} \nonumber\\
+\sum_{t=2}^{L_n}1\{ \pi_{t}^{0}=b,
b\notin
O^*(\underline{\underline{\theta}}), b \in O^*(\hat{\underline{\underline{\theta}}}^{t}),
u_{i}(\hat{\underline{\underline{\theta}}}^{t},\underline{\theta}_{i}^{'})=u_{\alpha^*}(\hat{\underline{\underline{\theta}}}^{t}), \nonumber\\
J_{i}(\hat{\underline{\underline{\theta}}}^{t},\epsilon)<\frac{\log
S_{\pi^0}(t-1)}{T_{\pi^0}^{i}(S_{\pi^0}(t-1))},
 J_{i}(\hat{\underline{\underline{\theta}}}^{t},\epsilon)\leq J_{i}(\underline{\underline{\theta}},\epsilon)-\delta \} \nonumber\\
\leq\sum_{t=2}^{L_n}1\{ \pi_{t}^{0}=b,
b\notin
O^*(\underline{\underline{\theta}}), b \in O^*(\hat{\underline{\underline{\theta}}}^{t}),
u_{i}(\hat{\underline{\underline{\theta}}}^{t},\underline{\theta}_{i}^{'})=u_{\alpha^*}(\hat{\underline{\underline{\theta}}}^{t}),
 T_{\pi^0}^{i}(S_{\pi^0}(t-1))<\frac{\log L_n}{J_{i}(\underline{\underline{\theta}},\epsilon)-\delta}\} \nonumber\\
+\sum_{t=2}^{L_n}1\{ \pi_{t}^{0}=b,
b\notin
O^*(\underline{\underline{\theta}}), b \in O^*(\hat{\underline{\underline{\theta}}}^{t}),
u_{i}(\hat{\underline{\underline{\theta}}}^{t},\underline{\theta}_{i}^{'})=u_{\alpha^*}(\hat{\underline{\underline{\theta}}}^{t}),
J_{i}(\hat{\underline{\underline{\theta}}}^{t},\epsilon)\leq
J_{i}(\underline{\underline{\theta}},\epsilon)-\delta \}. \nonumber
\end{gather}

Now, the first sum of the last inequality for $c=\frac{\log
L_n}{J_{i}(\underline{\underline{\theta}},\epsilon)-\delta}$   is equal to
\begin{eqnarray}
&&\sum_{t=2}^{L_n}1\{ \pi_{t}^{0}=b,
b\notin
O^*(\underline{\underline{\theta}}), b \in O^*(\hat{\underline{\underline{\theta}}}^{t}),u_{i}(\hat{\underline{\underline{\theta}}}^{t},\underline{\theta}_{i}^{'})=
 u_{\alpha^*}(\hat{\underline{\underline{\theta}}}^{t}), T_{\pi^0}^{i}(S_{\pi^0}(t-1))<c\}\nonumber\\
&\leq& \sum_{t=2}^{L_n}1\{ \pi_{t}^{0}=b,T_{\pi^0}^{i}(S_{\pi^0}(t-1))<c\}\nonumber\\
&=&\sum_{t=2}^{L_n}\sum_{s=0}^{\lfloor c/m_{i}^{b}\rfloor}1\{
\pi_{t}^{0}=
b,T_{\pi^0}^{i}(S_{\pi^0}(t-1))= s\, m_{i}^{b}+m^0_{i}\}\nonumber\\
&=&\sum_{s=0}^{\lfloor c/m_{i}^{b}\rfloor}\sum_{t=2}^{L_n}1\{
\pi_{t}^{0}=
b,T_{\pi^0}^{i}(S_{\pi^0}(t-1))= s\, m_{i}^{b}+m^0_{i}\}\nonumber\\
&\leq& \lfloor c/m_{i}^{b}\rfloor +1\nonumber\\
&\leq& \frac{c}{m_{i}^{b}} +1=\frac{\log L_n}{m_{i}^{b}(J_{i}
(\underline{\underline{\theta}},\epsilon)-\delta)}+1.\nonumber
\end{eqnarray}

Thus,
\begin{eqnarray}
&&E_{\underline{\underline{\theta}}}\sum_{t=2}^{L_n}1\{
\pi_{t}^{0}=b, b_0^t\in
O^*(\underline{\underline{\theta}}),
u_{i}(\hat{\underline{\underline{\theta}}}^{t},\underline{\theta}_{i}^{'})=u_{\alpha^*}(\hat{\underline{\underline{\theta}}}^{t}),
 T_{\pi^0}^{i}(S_{\pi^0}(t-1))<\frac{\log L_n}{J_{i}(\underline{\underline{\theta}},\epsilon)-\delta}\}\nonumber\\
&&\leq \frac{\log
L_n}{m_{i}^{b}(J_{i}(\underline{\underline{\theta}},\epsilon)-\delta)}+1.
\label{as8}
\end{eqnarray}

Furthermore,
\begin{eqnarray}
&&\sum_{t=2}^{L_n}1\{ \pi_{t}^{0}=b,
b\notin
O^*(\underline{\underline{\theta}}), b \in O^*(\hat{\underline{\underline{\theta}}}^{t}),u_{i}(\hat{\underline{\underline{\theta}}}^{t},\underline{\theta}_{i}^{'})=
u_{\alpha^*}(\hat{\underline{\underline{\theta}}}^{t}),
J_{i}(\hat{\underline{\underline{\theta}}}^{t},\epsilon)\leq J_{i}(\underline{\underline{\theta}},\epsilon)-\delta \}\nonumber\\
&&\leq \sum_{t=2}^{L_n}1\{
b\notin
O^*(\underline{\underline{\theta}}), b \in O^*(\hat{\underline{\underline{\theta}}}^{t}),
 J_{i}(\hat{\underline{\underline{\theta}}}^{t},\epsilon)\leq J_{i}(\underline{\underline{\theta}},\epsilon)-\delta \}\nonumber
\end{eqnarray}

Then from (C2) and Remark 3 we have that
\begin{eqnarray}
&&E_{\underline{\underline{\theta}}}\sum_{t=2}^{L_n}1\{
\pi_{t}^{0}=b, b\notin
O^*(\underline{\underline{\theta}}), b \in O^*(\hat{\underline{\underline{\theta}}}^{t}),u_{i}(\hat{\underline{\underline{\theta}}}^{t},
\underline{\theta}_{i}^{'})=u_{\alpha^*}(\hat{\underline{\underline{\theta}}}^{t}),
J_{i}(\hat{\underline{\underline{\theta}}}^{t},
\epsilon)\leq J_{i}(\underline{\underline{\theta}},\epsilon)-\delta \}\nonumber\\
&&\leq o(\log L_n). \label{as9}
\end{eqnarray}

Now we have that
$u_{i}(\hat{\underline{\underline{\theta}}}^{t},\underline{\theta}_{i}^{'})=
u_{\alpha^*}(\hat{\underline{\underline{\theta}}}^{t})>u_{s}(\hat{\underline{\underline{\theta}}}^{t},\underline{\theta}_{s}^{'})$
for any bandit $s$ which is contained in an optimal BFS of
$\underline{\underline{\theta}}$. Thus we can show the following inequalities
\begin{eqnarray}
&&\widetilde{T}_{\pi^0,1}^b(L_n)=\sum_{t=2}^{L_n}1\{ \pi_{t}^{0}=b,
b\notin
O^*(\underline{\underline{\theta}}), b \in O^*(\hat{\underline{\underline{\theta}}}^{t}),u_{i}(\hat{\underline{\underline{\theta}}}^{t},\underline{\theta}_{i}^{'})=
u_{\alpha^*}(\hat{\underline{\underline{\theta}}}^{t}),
u_{i}(\hat{\underline{\underline{\theta}}}^{t},
\underline{\theta}_{i}^{'})\leq z^*
(\underline{\underline{\theta}})-\epsilon \}
\nonumber\\
&&\leq\sum_{t=2}^{L_n}1\{
u_{s}(\hat{\underline{\underline{\theta}}}^{t},\underline{\theta}_{s}^{'})\leq
z^* (\underline{\underline{\theta}})-\epsilon \}
\nonumber\\
&&\leq\sum_{t=2}^{L_n}1\{
u_{s}(\hat{\underline{\underline{\theta}}}^{j},\underline{\theta}_{s}^{'})\leq
 z^* (\underline{\underline{\theta}})-\epsilon, \text{ for some }j\leq S_{\pi^0}(t-1) \}\nonumber\\
&&=\sum_{t=2}^{L_n}1\{
|\hat{\underline{\theta}}_{s}^{j}-\underline{\theta}_s|>\xi,\text{
for some }j\leq S_{\pi^0}(t-1) \}.\nonumber
\end{eqnarray}

Thus from condition (C3)
\begin{eqnarray}
&&E_{\underline{\underline{\theta}}}\sum_{t=2}^{L_n}1\{
\pi_{t}^{0}=b, b\notin
O^*(\underline{\underline{\theta}}), b \in O^*(\hat{\underline{\underline{\theta}}}^{t}),
u_{i}(\hat{\underline{\underline{\theta}}}^{t},\underline{\theta}_{i}^{'})=
u_{\alpha^*}(\hat{\underline{\underline{\theta}}}^{t}),
u_{i}(\hat{\underline{\underline{\theta}}}^{t},\underline{\theta}_{i}^{'})\leq z^* (\underline{\underline{\theta}})-\epsilon \}\nonumber\\
&&\leq o(\log L_n). \label{as10}
\end{eqnarray}

Finally, it follows from \eqref{as8}, \eqref{as9} and \eqref{as10}
that
\begin{equation}
E_{\underline{\underline{\theta}}}\widetilde{T}_{\pi^0}^{b}(L_n)\leq
 \frac{\log L_n}{m_{i}^{b}(J_{i}(\underline{\underline{\theta}},\epsilon)-\delta)} +1 +o(\log L_n)+o(\log L_n).\nonumber
\end{equation}

Now from the definition of
$J_{i}(\underline{\underline{\theta}},\epsilon)$ and (C1) we have
that
\begin{equation}
\lim_{\epsilon\rightarrow
0}J_{i}(\underline{\underline{\theta}},\epsilon)=
K_i(\underline{\underline{\theta}}),\text{ for }i\in
D(\underline{\underline{\theta}})\text{  and  }
\lim_{\epsilon\rightarrow
0}J_{i}(\underline{\underline{\theta}},\epsilon)=\infty,\text{ for
}i\notin D(\underline{\underline{\theta}}).\nonumber
\end{equation}

Thus
\begin{eqnarray}
&&\uplim_{n\rightarrow
\infty}\frac{E_{\underline{\underline{\theta}}}\widetilde{T}_{\pi^0,2}^{b}(L_n)}{\log
L_n} \leq\frac{1}{m_{i}^{b}\,K_{i}(\underline{\underline{\theta}})}, \text{ for
all } i\in D(\underline{\underline{\theta}}), i\in b, b\notin
O^*(\underline{\underline{\theta}})
,\nonumber\\
&&\uplim_{n\rightarrow
\infty}\frac{E_{\underline{\underline{\theta}}}\widetilde{T}_{\pi^0,2}^{b}(L_n)}{\log
L_n}=0, \text{ for all } i\notin
D(\underline{\underline{\theta}}),i\in b, b\notin
O^*(\underline{\underline{\theta}}) \text{ and}\nonumber
\end{eqnarray}\\
\begin{equation}
\uplim_{n\rightarrow
\infty}\frac{E_{\underline{\underline{\theta}}}\widetilde{T}_{\pi^0,1}^b(L_n)}{\log
L_n}=0, \text{ for all } b\notin
O^*(\underline{\underline{\theta}}).\nonumber
\end{equation}
This completes the proof.%$\blacksquare$
\endproof

\end{appendix}
%
%   or
%
% \begin{APPENDICES}
% \section{<Title of Section A>}
% \section{<Title of Section B>}
% etc
% \end{APPENDICES}

% Acknowledgments here
{\bf Acknowledgments}{ We acknowledge support for this work from the National Science Foundation, NSF grant CMMI-16-62629.}

% References here (outcomment the appropriate case)

% CASE 1: BiBTeX used to constantly update the references
%   (while the paper is being written).
%\bibliographystyle{informs2014} % outcomment this and next line in Case 1
\bibliography{mab2018} % if more than one, comma separated

\end{document}